\documentclass[sigconf, screen, nonacm]{acmart}
\usepackage{graphicx}
\usepackage{enumitem}
\usepackage{comment}
\usepackage{booktabs}
\usepackage{multirow}
\usepackage{geometry}
\PassOptionsToPackage{table}{xcolor}
\usepackage{multirow}
\usepackage{colortbl}
\usepackage{tikz}
\usepackage{pgfplots}
\pgfplotsset{compat=1.18}
\definecolor{color_zeroshot}{RGB}{220,210,195} 
\definecolor{color_sft}{RGB}{189,175,162}     
\definecolor{color_naive}{RGB}{154,175,169}   
\definecolor{color_agent}{RGB}{110,127,136}   
\usepackage{pifont}
\usepackage{colortbl}   
\usepackage{xcolor}     
\definecolor{LightCyan}{rgb}{0.85,0.93,0.97}
\usepackage{tabularx}
\usepackage{subcaption}

\begin{document}

\title[DynFlux]{DynFlux: Learning to Dynamically Balance Consensus and Conflict in Multimodal Emotion Recognition}

\author{Zeheng Wang}
\affiliation{%
  \institution{Northeast Normal University}
  \country{China}}
\email{wangzeheng624@nenu.edu.cn}

\author{Hao Zhang}
\affiliation{%
  \institution{University of the Chinese Academy of Sciences}
  \country{China}}

\author{Yige Wang}
\affiliation{%
  \institution{Boston University}
  \country{USA}}

\author{Xi Xiao}
\affiliation{%
  \institution{University of Alabama at Birmingham}
  \country{USA}}

\author{Hao Xu}
\affiliation{%
  \institution{Harvard University}
  \country{USA}}
\email{haxu@bwh.harvard.edu}

\renewcommand{\shortauthors}{Wang et al.}

\begin{abstract}
Multimodal emotion recognition remains challenging due to the intrinsic heterogeneity of different modalities. Effectively leveraging both cross-modal consensus (consistency) and conflict (inconsistency) is essential, as they jointly provide complementary cues that enhance emotion understanding from diverse perspectives. 
Traditional multimodal methods often suffer from semantic ambiguity and rarely explicitly model the interplay between cross-modal consensus and conflict. Moreover, existing approaches struggle to capture and exploit the dynamic interactions between these two complementary factors. To overcome these limitations, we propose DynFlux, a model guided by a temporal heterogeneous graph that encodes both consensus and conflict relationships across modalities. DynFlux introduces a dedicated graph branch that explicitly captures cross-modal dependencies through sparse temporal graphs, and modulates the conventional multimodal emotion recognition pipeline at three hierarchical stages: adaptive feature modulation during feature extraction, guided aggregation modulation during feature integration, and fusion strategy modulation at the multimodal fusion stage. Through this design, DynFlux dynamically balances cross-modal consensus and conflict, enabling richer and more robust emotion understanding. Extensive experiments on benchmark datasets confirm that DynFlux consistently outperforms state-of-the-art methods, and its generality is further validated on multimodal intent recognition and multimodal sarcasm detection tasks.
\end{abstract}



\maketitle

\section{Introduction}
Multimodal Emotion Recognition (MER) aims to understand human emotions in videos by integrating time-series data from multiple modalities including visual, acoustic, and linguistic information, establishing itself as a central research area in affective computing~\cite{Liang2021Attention,Lv2021PMR,Poria2023Iceberg}. Compared to unimodal approaches, multimodal data provides complementary and enriched emotional representations, enabling more accurate characterization of human affective states. With recent advances in deep learning techniques, MER research has rapidly evolved and demonstrated extensive practical value across diverse domains such as intelligent education~\cite{Petrovica2017ATS}, product evaluation~\cite{Melville2009Blogs}, and robotics~\cite{Liu2017FEERHRI}.

\begin{figure}[!t] 
    \centering
    \includegraphics[width=\columnwidth, height=0.35\textheight, keepaspectratio]{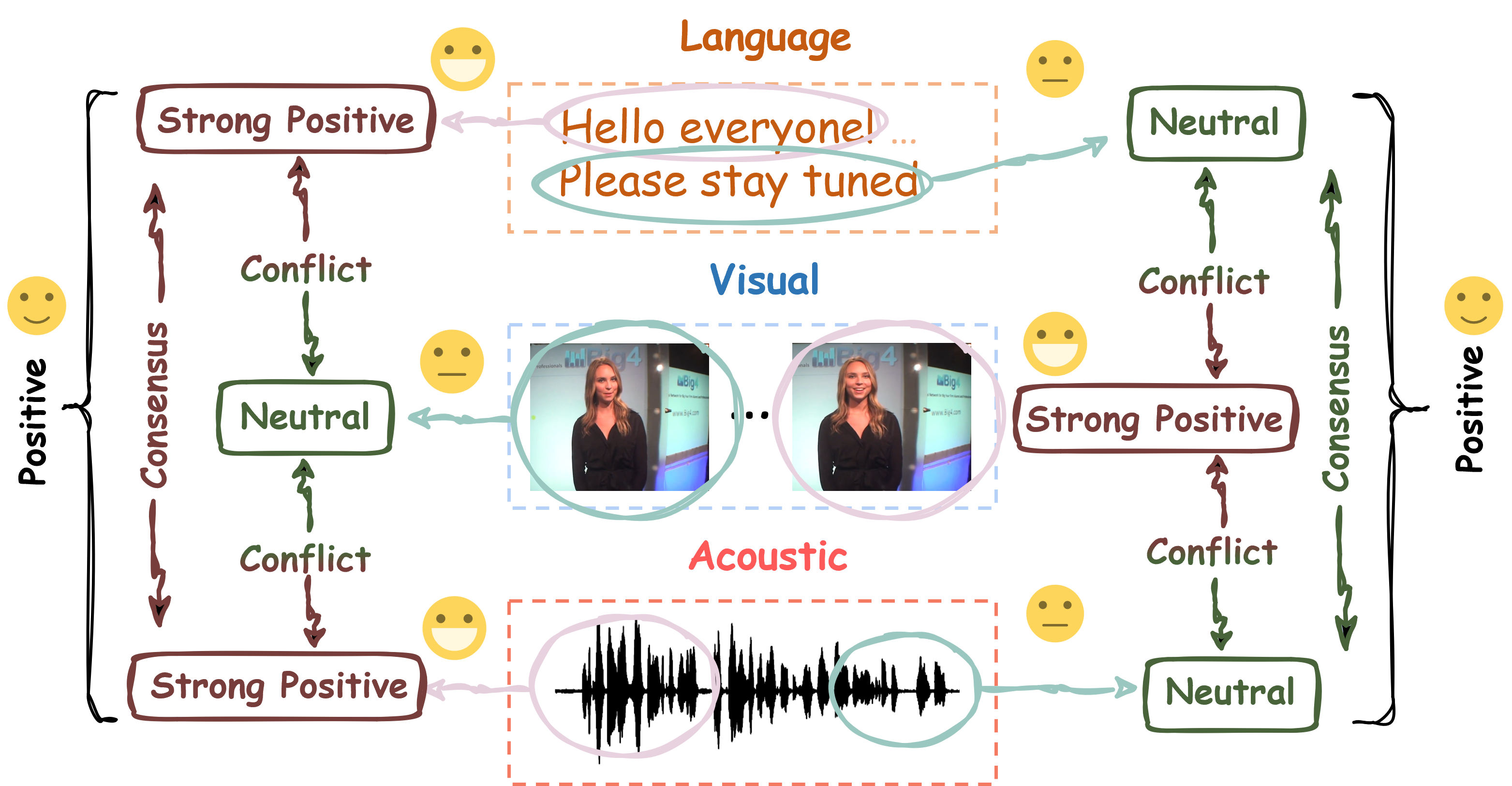}
    \caption{ Example of dynamic consensus and conflict relationships across modalities in multimodal emotion recognition. 
    The language modality expresses a strong positive sentiment, while the acoustic signal becomes neutral in later segments and the visual cues remain relatively stable. 
    This temporal inconsistency leads to alternating consensus and conflict among modalities. 
    The observation reveals that cross-modal agreement varies over time, motivating our framework to explicitly model and dynamically balance both consensus and conflict signals.}
    \label{fig:intro}
\end{figure}

In Multimodal Emotion Recognition (MER), different modalities within the same video segment often complement each other, exhibiting both semantic and emotional consensus and conflict. Although extensive research has been conducted in the field of MER~\cite{Zadeh2017TFN,Hazarika2020MISA,Tsai2019MulT,Shi2025GSCon,Huan2024TriSAT}, existing approaches tend to either directly fuse multimodal representations or distill conflict before fusion. Their core idea is to eliminate discrepancies while preserving consensus, regarding divergent information as detrimental to both the model optimization process and final predictions. Recent studies have conducted in-depth analyses of inter-modal discrepancies, arguing that outcome biases arise from the dominance of modalities with superior representational capacity~\cite{Zhong2025CMC,Fang2025EMOE,Shi2025GSCon}.

We argue that in multimodal emotion recognition, both consensus and conflict information across modalities play crucial roles. These elements do not exist in isolation but dynamically interact to construct emotional representations: consensus reinforces the shared semantic core, while conflict injects complementary perspectives. Together, they evolve along temporal sequences to form a robust emotional understanding process. For example, as shown in Fig.~\ref{fig:intro}, multimodal emotional cues exhibit dynamic patterns of consensus and conflict.

\begin{figure*}[t]
    \centering
    \includegraphics[width=\linewidth]{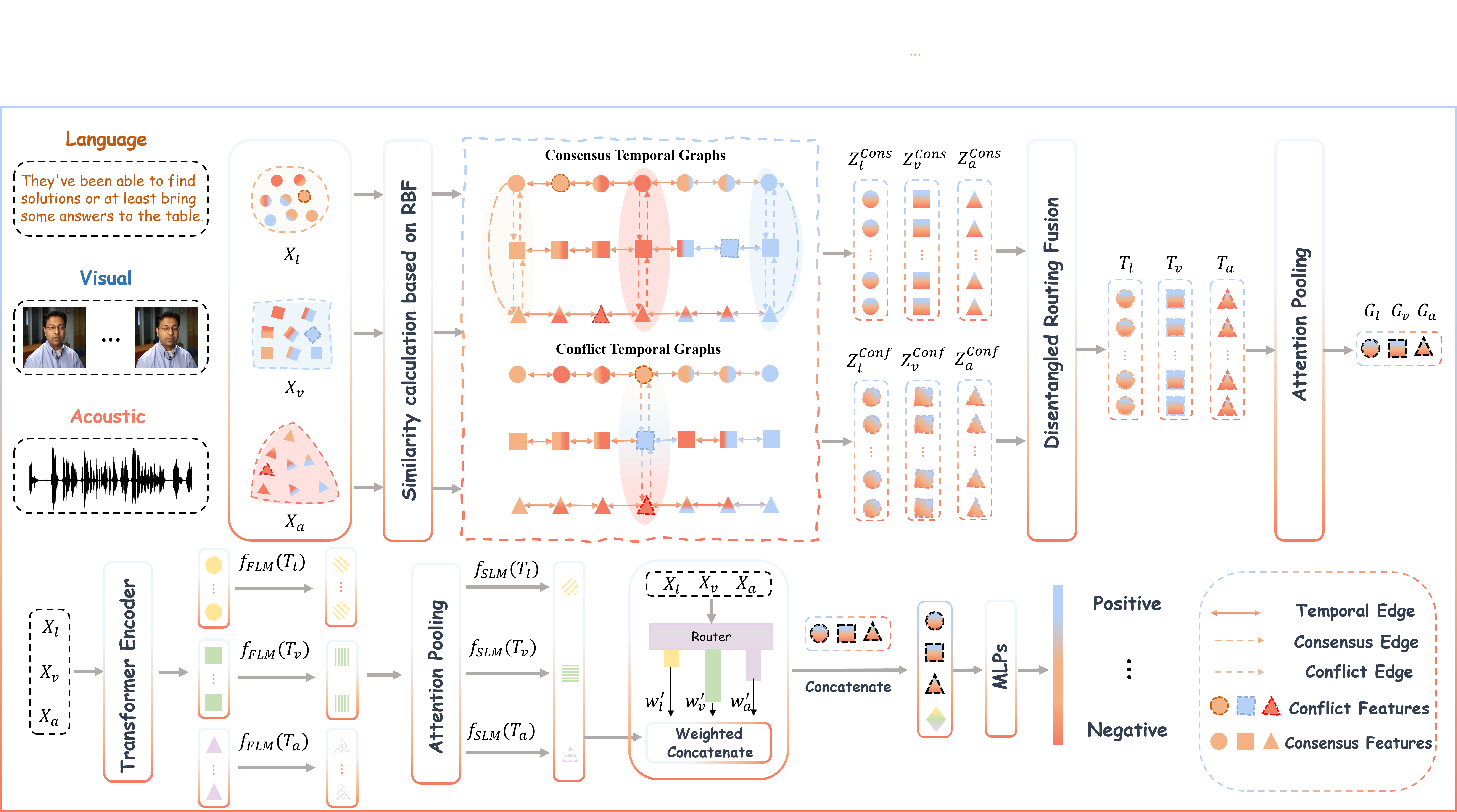}
    \caption{The framework of the DynFlux. Given the input multimodal sequences $\{X_l, X_v, X_a\}$, DynFlux first uses Transformer encoders to extract modality-specific temporal features for language, visual, and acoustic streams (Sec.~\ref{sec:preliminaries}). In the graph branch, these features first undergo Radial Basis Function (RBF) based cross-modal consensus and divergence computation(Sec.~\ref{sec:cons_conf}), and used to construct sparse temporal heterogeneous graphs (Sec.~\ref{sec:graphs}). A spatial–temporal block performs heterogeneous graph attention and temporal self-attention on these graphs to obtain consensus features $Z^{\mathrm{cons}}_m$ and conflict features $Z^{\mathrm{conf}}_m$. Through the Disentangled Routing Fusion (DRF) module, DynFlux fuses $Z^{\mathrm{cons}}_m$ and $Z^{\mathrm{conf}}_m$ into time-varying modulation signals $T_m$, which guide hierarchical interventions at the feature, aggregation, and fusion stages (Sec.~\ref{sec:hier_mod}). Finally, the modulated unimodal representations and graph features are fused to produce the final prediction.}
    \label{fig:framework}
\end{figure*}

To address these challenges and leverage evolving cross-modal interactions, we propose the DynFlux framework: an approach for multimodal emotion recognition that dynamically balances cross-modal consensus and conflict through a temporal heterogeneous graph. As illustrated in Figure~\ref{fig:framework}, the framework projects features from different modalities into consensus and conflict subspaces respectively. Subsequently, the Radial Basis Function (RBF) kernel is employed to quantify the degrees of consensus and conflict between different modalities. At the same time, we use the Maximum Mean Discrepancy (MMD) to constrain the consensus and conflict spaces, so that the feature consensus and conflict projected into these spaces are maximized, thereby enhancing the distinction between consensus and conflict representations. Through this methodology, we construct a weighted graph structure that distinguishes consensus edges (strong consistency) from conflict edges (strong inconsistency), while incorporating bidirectional temporal chains to preserve sequential continuity.

The representations extracted by the heterogeneous graph encode both consensus and conflict. To emphasize these critical signals, we initially modeled only the most salient fine grained features of consensus and conflict, which left cross modal interactions underutilized. To fully leverage the extracted information, we introduce a modulation mechanism. By performing content aware routing between the consensus and conflict embeddings, we generate time varying modulation signals that enable hierarchical interventions. Specifically, we apply adaptive scaling during feature extraction to amplify modality specific cues, refine attention during temporal aggregation to highlight salient segments, and strategically modulate modality weights during the final fusion. This architecture captures the evolving dynamics between consensus and conflict and integrates seamlessly into standard multimodal emotion recognition (MER) pipelines. Given the similarity of the tasks, we further transfer the method to multimodal intent recognition (MIR) and multimodal sarcasm detection (MSD), where we observe consistent gains. The main contributions of this work can be summarized as follows:
\begin{itemize}[leftmargin=*]
\item 
We propose a temporal heterogeneous graph framework designed to capture dynamic cross-modal consensus and conflict. By explicitly modeling and distilling these interactions, it enables more nuanced and effective multimodal understanding

\item
To overcome the limitations of static fusion, we design a content-aware dynamic routing module that adaptively balances consensus and conflict streams. The generated modulation signals guide hierarchical interventions at the feature, aggregation, and fusion levels.

\item 
We conduct extensive experiments on CMU-MOSI~\cite{Zadeh2016MOSI}, CMU-MOSEI~\cite{Zadeh2018MOSEI}, MIntRec~\cite{Zhang2022MIntRec}, and MMSD~\cite{cai2019mmsd} datasets, achieving superior or comparable results to state-of-the-art methods. We further extend DynFlux from multimodal emotion recognition (MER) to intent recognition (MIR) and sarcasm detection (MSD), demonstrating its generality and robustness.
\end{itemize}

\section{Related Works}
\subsection{Multimodal Emotion Recognition}
 Multimodal Emotion Recognition (MER) aims to integrate multimodal information to determine human emotions in videos. Different modalities convey emotions from diverse perspectives, and effectively integrating multiple modalities is key to enhancing multimodal emotion recognition. There are two mainstream approaches in MER: multimodal fusion methods~\cite{Zadeh2017TFN,zadeh2018mfn,liu2018lmf,morency2011towards,peng2022bml} and cross-modal learning methods~\cite{Liang2021Attention,Hazarika2020MISA,Tsai2019MulT,Lv2021PMR,Fang2025EMOE,li2023dmd}.

 The first research direction aims to integrate multimodal features more effectively. Early multimodal fusion approaches primarily employed simple concatenation operations~\cite{morency2011towards}. Subsequently, TFN~\cite{Zadeh2017TFN} introduced a progressive fusion strategy. Further advancing this line of work, studies such as Hazarika et al.~\cite{tsai2019lfmr,Hazarika2020MISA,yang2022drler} decoupled multimodal features into modality-invariant and modality-specific components. DMD~\cite{li2023dmd} extended this by performing distillation on the decoupled features, while EMOE~\cite{Fang2025EMOE} and GSCon~\cite{Shi2025GSCon} delved deeper into balancing different modalities within the fused representations. 

 In parallel, cross-modal learning focuses on enhancing interactions among multiple modalities. For instance, MulT~\cite{Tsai2019MulT} proposed a multimodal transformer incorporating a cross-modal attention mechanism, which effectively captures latent adaptations and correlations across modalities. Recent research has also leveraged large multimodal language models to facilitate cross-modal interactions~\cite{wu2024nextgpt,wang2024wisdom}, including unified emotional understanding and generation~\cite{zhu2025emosym,zhu2026uniemo,11658652}, retrieval-augmented and agent-based emotion recognition~\cite{wang2026affectagent,wang2026navigating}, micro-expression and micro-gesture analysis~\cite{liu2026aullm++,wang2026micro,10.1145/3767308.3835254,10.1007/978-981-92-4529-1_3}, and change captioning~\cite{wang2026rsiccllm,wang2026dfm}. Related multimodal modeling advances span vision-language-action models~\cite{zhu2026h,zhu2026delta,zhu2026ati,zhu2026learning}, face anti-spoofing and forgery detection~\cite{ma2026cm,ma2026clip,ma2026flexible,tan2026veritas,tan2026videoveritas,shi2026hydraprompt}, security of generative vision-language models~\cite{DBLP:conf/iclr/PanCWDW26,DBLP:journals/corr/abs-2605-07399}, signed-graph structural modeling~\cite{gao2026polarityasymmetricstructuralcalibrationlink,wang2026signedgraph}, and speech synthesis~\cite{wang2027manchutts}.

 In contrast to existing MER methods that primarily decouple modalities, our framework directly addresses the extraction and explicit modeling of cross-modal consensus and conflict.

 \subsection{Multimodal Consensus and Conflict Modeling}
Modeling consensus and conflict has become central to reliable multimodal learning, evolving from specialized concepts to a mainstream paradigm. Methods such as C3N~\cite{shi2023c3n} inject cross-modal consensus into local representations after global alignment, while GCCL~\cite{dai2024gccl} enforces emotion-level consistency via graph-structured consensus units. STiL~\cite{du2025stil} further employs classifier-aggregated pseudo-labels with prototype smoothing in semi-supervised tabular-image learning.

In contrast, disagreement-aware approaches explicitly leverage inter-modal conflicts. Uni-X~\cite{hao2025unix} mitigates gradient-level clashes through layer-wise segregation of transformer modules. CoDAF~\cite{liu2025codaf} resolves spatial and semantic misalignment in RGB-IR UAV imagery via offset-guided fusion. Mixed Signals~\cite{srikanth2025mixed} correlates prediction divergence with annotator uncertainty in empathy detection. 

Building upon these advances, the natural progression is to jointly leverage both consensus and conflict~ \cite{karri2025ugcemt,tan2025dpa,cohn2024mmla_review,sun2024redcore}. MCAN~\cite{gao2025mcan} disentangles and regularizes alignment-conflict components across both representation and output levels. RAVEN~\cite{shi2025raven} computes token-level consensus via query-conditioned gating and refines high-disagreement samples for robust multimodal QA. Existing research reveals a notable gap: few methods explicitly capture and effectively utilize the dynamic consensus and conflict inherent in different modalities.

\section{Method}

\subsection{Preliminaries}
\label{sec:preliminaries}
As in most previous emotion recognition tasks, our primary objective is to perform sentiment classification or regression for each multimodal sample using three modalities: language (\(l\)), vision (\(v\)), and audio (\(a\)). For each utterance, the feature sequence of modality \( m \in \{l, v, a\} \) is represented as a matrix $\mathbf{X}_m =
[\mathbf{x}_m^{(1)}, \mathbf{x}_m^{(2)}, \dots, \mathbf{x}_m^{(T_m)}]^{\top}
\in \mathbb{R}^{T_m \times d_m}$, where \( T_m \) denotes the temporal length of the sequence (i.e., the number of time steps, such as tokens for language or frames for vision), and \( d_m \) denotes the feature dimensionality. As illustrated in Figure~1, we propose a multimodal emotion recognition framework that is guided by cross-modal temporal consensus and conflict information.

\subsection{Multimodal Consensus-Conflict Computation}
\label{sec:cons_conf}
We construct a heterogeneous graph by quantifying cross-modal consensus and conflict relationships within a unified framework. To explicitly measure consensus and conflict between modalities, we propose a set of complementary metrics based on the Radial Basis Function (RBF) kernel. By leveraging the properties of a high-dimensional Reproducing Kernel Hilbert Space (RKHS), we project the fine-grained features of each modality into this high-dimensional space, enabling the capture of more nuanced consensus and conflict. Meanwhile, the computed values are employed as edge weights in the graph neural network.

For each directed modality pair \(m \to n\) (where \(m, n \in \{l, v, a\}\) and \(m \neq n\)), we compute the following, we define \(K(\cdot,\cdot)\) as the function that computes the pairwise RBF kernel similarity between all tokens in two input feature sequences.
\begin{equation}
k(\mathbf{x}, \mathbf{y}) = \exp\!\left(-\frac{\|\mathbf{x} - \mathbf{y}\|_2^2}{2\sigma^2}\right).
\end{equation}
For the consensus score matrix \(\mathbf{S}_{m\to n}^{\text{cons}} \in \mathbb{R}^{T_m \times T_n}\), we first project both the source modality \(\mathbf{X}_m\) and target \(\mathbf{X}_n\) modality into a consensus semantic space, and then compute the RBF similarity:
\begin{equation}
\mathbf{S}_{m\to n}^{\text{cons}}[i,j] = k(\mathbf{W}_{m \to n}^{\text{cons}}(\mathbf{x}_m^{(i)}), \mathbf{W}_{m\to n}^{\text{cons}}(\mathbf{x}_n^{(j)})).
\end{equation}
Similar to the consensus score matrix \(\mathbf{S}_{m\to n}^{\text{conf}} \in \mathbb{R}^{T_m \times T_n}\), we project both the source and target modalities into a conflict semantic space before computing the RBF similarity:
\begin{equation}
\mathbf{S}_{m\to n}^{\text{conf}}[i,j] = 1 - k(\mathbf{W}_{m \to n}^{\text{conf}}(\mathbf{x}_m^{(i)}), \mathbf{W}_{m \to n}^{\text{conf}}(\mathbf{x}_n^{(j)})).
\end{equation}
Here, \(\mathbf{W}_{m\to n}^{\text{cons}}\) and \(\mathbf{W}_{m\to n}^{\text{conf}}\) are the learnable projection matrices for consensus and divergence modeling, respectively. To ensure the consensus and conflict projections map the original features into authentic consensus and divergence spaces, we employ Maximum Mean Discrepancy (MMD) to quantify the deviations after projection,The general form of the MMD formula is expressed as:
\begin{equation}
\begin{aligned}
\text{MMD}^2(P, Q) &= 
\mathbb{E}_{x,x' \sim P}[k(x, x')] 
+ \mathbb{E}_{y,y' \sim Q}[k(y, y')] \\
&\quad - 2\,\mathbb{E}_{x \sim P, y \sim Q}[k(x, y)]
\end{aligned}
\end{equation}
We utilize this formula to compute the losses for consensus and conflict projections:
\begin{equation}
\mathcal{L}_\text{cons}=\text{MMD}^2(\mathbf{W}_{m \to n}^{\text{cons}}(\mathbf{X}_m),\mathbf{W}_{m \to n}^{\text{cons}}(\mathbf{X}_n))
\end{equation}
\begin{equation}
\mathcal{L}_\text{conf}=-\text{MMD}^2(\mathbf{W}_{m \to n}^{\text{conf}}(\mathbf{X}_m),\mathbf{W}_{m \to n}^{\text{conf}}(\mathbf{X}_n))
\end{equation}

\subsection{Multimodal Temporal Graphs}
\label{sec:graphs}
We construct a heterogeneous multimodal temporal graph to extract and enhance consensus and conflict information.
\paragraph{Node Definition.}
We consider three modalities: language (\(l\)), vision (\(v\)), and audio (\(a\)). Each produces a sequence of feature tokens that serve as graph nodes:
\begin{equation}
\mathcal{V} =
\bigcup_{m\in\{l,v,a\}}
\{v_m^{(t)} \mid \mathbf{x}_m^{(t)},~ t=1,\dots,T_m\}.
\end{equation}
\paragraph{Edge Definition.}
To manage computational complexity and mitigate noise from a fully connected graph, we employ a sparse edge generation strategy. For each directed modality pair \(m \to n\) (where \(m, n \in \{l, v, a\}\) and \(m \neq n\)), we construct edges using the consensus and conflict score matrices $\mathbf{S}_{m\to n}^{\text{cons}}$ and $\mathbf{S}_{m\to n}^{\text{conf}}$. Let $i \in \{1, \dots, T_m\}$ index a node $v_m^{(i)}$ in the source modality $m$, and $j \in \{1, \dots, T_n\}$ index a node $v_n^{(j)}$ in the target modality $n$. Each entry $\mathbf{S}_{m\to n}^{(\cdot)}[i,j]$ measures the semantic relationship between
$v_m^{(i)}$ and $v_n^{(j)}$, where $(\cdot)\!\in\!\{\text{cons}, \text{conf}\}$.
A directed edge $(v_m^{(i)},v_n^{(j)})$ is added if $v_n^{(j)}$ is among the Top-K most similar
targets of $v_m^{(i)}$ and the similarity score exceeds the threshold $\tau$.
The resulting cross-modal edge sets are:
\begin{equation}
\resizebox{\linewidth}{!}{$
\begin{aligned}
\mathcal{E}^{\text{cons}} &= 
\bigcup_{m \to n} \bigcup_{i=1}^{T_m} 
\{(v_m^{(i)},v_n^{(j)}) ~|~ 
j \in \text{TopK}_i(\mathbf{S}_{m\to n}^{\text{cons}}),
~\mathbf{S}_{m\to n}^{\text{cons}}[i,j]>\tau \}, \\
\mathcal{E}^{\text{conf}} &= 
\bigcup_{m \to n} \bigcup_{i=1}^{T_m} 
\{(v_m^{(i)},v_n^{(j)}) ~|~ 
j \in \text{TopK}_i(\mathbf{S}_{m\to n}^{\text{conf}}),
~\mathbf{S}_{m\to n}^{\text{conf}}[i,j]>\tau \}.
\end{aligned}
$}
\end{equation}

\paragraph{Temporal Continuity.}
To capture dynamic consensus and conflict, we further incorporate temporal modeling. Within each modality, temporal order is maintained through bidirectional sequential edges:
\begin{equation}
\resizebox{\linewidth}{!}{$\mathcal{E}^{\text{temp}} =
\bigcup_{m\in\{l,v,a\}}
\{(v_m^{(t)},v_m^{(t+1)}), (v_m^{(t+1)},v_m^{(t)}) \mid t=1,\dots,T_m-1\}$}.
\end{equation}

\paragraph{Final Graph Assembly.}
At the final stage, we assemble two complementary heterogeneous graphs:
\begin{equation}
\mathcal{G}_{\text{cons}} = (\mathcal{V},\, \mathcal{E}^{\text{cons}} \cup \mathcal{E}^{\text{temp}}),
\end{equation}
\begin{equation}
\mathcal{G}_{\text{conf}} = (\mathcal{V},\, \mathcal{E}^{\text{conf}} \cup \mathcal{E}^{\text{temp}}).
\end{equation}
Where $\mathcal{G}_{\text{cons}}$ captures cross-modal consensus and $\mathcal{G}_{\text{conf}}$ highlights regions of conflict.

\paragraph{Heterogeneous Message Passing}
Upon graph construction, we proceed to message passing between nodes. The message passing is performed through a \textbf{Spatial–Temporal Block} 
$\Phi_{\text{ST}}$, which sequentially applies a spatial aggregation step followed by a temporal refinement step. 
Formally, given node representations propagated over a heterogeneous graph $\mathcal{G}$, 
the block operates as:
\begin{equation}
\Phi_{\text{ST}}(\cdot, \mathcal{G}) 
= \mathrm{H\text{-}GAT}(\cdot, \mathcal{G}) 
+ \mathrm{MHA}\big(\mathrm{H\text{-}GAT}(\cdot, \mathcal{G})\big),
\end{equation}
where $\mathrm{H\text{-}GAT}$ denotes heterogeneous graph attention, 
and $\mathrm{MHA}$ performs multi-head self-attention along the temporal dimension.

The final consensus and conflict representations, \(\mathbf{Z}^{\text{cons}}\) and \(\mathbf{Z}^{\text{conf}}\), are then produced by applying this block to their respective inputs:
\begin{align}
\mathbf{Z}^{\text{cons}}_m &= \Phi_{\text{ST}}(\mathbf{X}_{m}, \mathcal{G}_{\text{cons}}), \\
\mathbf{Z}^{\text{conf}}_m &= \Phi_{\text{ST}}(\mathbf{X}_{m}, \mathcal{G}_{\text{conf}}).
\end{align}

\subsection{Consensus-Conflict Hierarchical Modulation}
\label{sec:hier_mod}
\paragraph{Signal Generation.}
We generate fine-grained modulation features \(\mathbf{T}_m \in \mathbb{R}^{T_m \times D_g}\), by dynamically fusing consensus embeddings $\mathbf{Z}^{\text{cons}}_m$ and conflict embeddings $\mathbf{Z}^{\text{conf}}_m$ through our \textbf{Disentangled Routing Fusion (DRF)} mechanism. DRF replaces the simple averaging approach that tends to cause information cancellation with a learnable router, which performs content-aware dynamic soft-selection between the two information streams at each time step.
The core of DRF is a router MLP, \(\phi_{\text{router}}\), which takes the concatenated consensus and conflict features at each time step \(t\) as input to produce a scalar routing weight \(w_m(t)\):
\begin{equation}
w_m(t) = \phi_{\text{router}}\left( \left[ \mathbf{Z}^{\text{cons}}_m[t] \,;\, \mathbf{Z}^{\text{conf}}_m[t] \right] \right).
\end{equation}
where \([\cdot \,;\cdot]\) denotes concatenation. The final intervention signal \(\mathbf{T}_m\) is then formed by a linear interpolation between the two streams, guided by this dynamic weight:
\begin{equation}
\mathbf{T}_{m}[t] = w_m(t)\,\mathbf{Z}^{\text{cons}}_{m}[t] + (1 - w_m(t))\,\mathbf{Z}^{\text{conf}}_{m}[t].
\end{equation}
The resulting tensor \(\mathbf{T}_m\) provides time-varying guidance that captures the evolving, learned balance between cross-modal agreement and conflict, rather than relying on a fixed heuristic like feature norm.

\paragraph{Frame-Level Modulation.}
To prevent the averaging out of critical information before feature aggregation and leverage cross-modal consensus-conflict patterns for fine-grained consensus and conflict reinforcement, we adaptively modulate unimodal features at each temporal step. We employ a Transformer encoder for feature extraction and modulate its output $\{\mathbf{C}_{m}\}$ using the obtained $\mathbf{T}_m$ through the following process:

\begin{equation}
[\gamma^{\text{flm}}_t, \beta^{\text{flm}}_t, g^{\text{flm}}_t] = \phi_{\text{flm}}(\mathbf{T}_m[t]).
\end{equation}
where $\phi_{\text{flm}}$ is the temporal feature modulation network that maps $\mathbf{T}_m[t]$ to modulation parameters: $\gamma^{\text{flm}}_t$ for scaling, $\beta^{\text{flm}}_t$ for shifting, and $g^{\text{flm}}_t$ for gating.

\begin{equation}
\mathbf{m}^{\text{flm}}_t = (\tanh(\gamma^{\text{flm}}_t) + 1) \odot \mathbf{C}_m[t] + \lambda_1 \cdot \tanh(\beta^{\text{flm}}_t).
\end{equation}
where $\mathbf{m}^{\text{flm}}_t$ is the temporal feature modulation component combining adaptive scaling and constrained translation, , $\lambda_1$ controls the strength of shifting.

\begin{equation}
\mathbf{C}'_m[t] = \mathbf{C}_m[t] + \sigma(g^{\text{flm}}_t) \odot \mathbf{m}^{\text{flm}}_t.
\end{equation}
where the gated residual connection preserves original features while allowing controlled modulation through $\sigma(g^{\text{flm}}_t)$.

\paragraph{Sequence-Level Modulation}
To prevent interference from irrelevant or conflicting temporal segments during feature aggregation and leverage cross-modal consensus-conflict dynamics for dynamic temporal importance adjustment, we introduce Aggregation Modulation to refine attention pooling:

\begin{equation}
[\Delta e^{\text{slm}}_t, g^{\text{slm}}_t] = \phi_{\text{slm}}(\mathbf{T}_m[t]).
\end{equation}
where $\phi_{\text{slm}}$ is the aggregation modulation network that generates attention adjustment $\Delta e^{\text{slm}}_t$ and gate $g^{\text{slm}}_t$ from $\mathbf{T}_m[t]$.

\begin{equation}
E'_t = E_t + \sigma(g^{\text{slm}}_t) \cdot \Delta e^{\text{slm}}_t.
\end{equation}
where the attention score modulation follows a similar gated residual design as Temporal Feature Modulation, with $E_t$ being the original attention score.

\begin{equation}
\alpha_t = \frac{\exp(E'_t)}{\sum_{j=1}^{T}\exp(E'_j)},
\end{equation}

\begin{equation}
\mathbf{H}_m = \sum_{t=1}^{T} \alpha_t \cdot \mathbf{C}'_m[t].
\end{equation}
where the modulated attention scores $\{E'_t\}$ are normalized via softmax to produce attention weights $\alpha_t$, which are then used to aggregate the temporally modulated features $\mathbf{C}'_m$, yielding the pooled representation $\mathbf{H}_m$ for each modality $m \in \{l,a,v\}$.
\paragraph{Cross-Modal Modulation.}
We employ a router to generate modality weights, fusing multimodal features. Specifically, we concatenate the original features $\mathbf{X}_{m}$ and feed them into the router network, which outputs the corresponding weights $\mathbf{w}_{m}$ for the three modalities. To address the bias in router weight allocation when conflicts exist between modalities and adjust the fusion strategy based on global consensus-conflict patterns, we introduce Fusion Modulation:
\begin{equation}
\mathbf{f}_m = \text{AttenPool}(\mathbf{T}_m).
\end{equation}
where $\mathbf{f}_m \in \mathbb{R}^{D_g}$ is the global feature obtained by applying attention pooling to the $\mathbf{T}_m$. The implementation of attention pooling follows the formulation defined in the Aggregation Modulation section

\begin{equation}
[\Delta w^{\text{cmm}}_m, g^{\text{cmm}}_m] = \phi_{\text{cmm}}(\mathbf{f}_m).
\end{equation}
where $\phi_{\text{cmm}}$ is the fusion modulation network that generates fusion weight adjustment $\Delta w^{\text{cmm}}_m$ and gate $g^{\text{cmm}}_m$ from the global feature.

\begin{equation}
w_m' = w_m + \sigma(g^{\text{cmm}}_m) \cdot \tanh(\Delta w^{\text{cmm}}_m).
\end{equation}
Subsequently, the obtained $w'_m$ is used for multimodal weighted fusion.

\subsection{Final Prediction}
The final prediction is obtained through a hierarchical fusion process. First, the modulated unimodal representations \(\mathbf{H}_m\) are fused according to the strategic weights \(w'_m\). Concurrently, the graph branch outputs \(\mathbf{Z}^{\text{cons}}_m\) and \(\mathbf{Z}^{\text{conf}}_m\) are aggregated into vectors along the temporal dimension through attention pooling. We use the same method as previously described in Modulation Generation to fuse the consensus and conflict information, obtaining \(\mathbf{G}_m\), and then concatenate the $\mathbf{G}_m$ from different modalities to obtain G. The final fused representation is:
\begin{equation}
y_{\text{final}} = \mathrm{MLP}\left( \left[ \sum_{m \in \{l,a,v\}} w'_m \cdot \mathbf{H}_m \,;\, \mathbf{G} \right] \right).
\end{equation}

\subsection{Objective Optimization}
Based on the two categories of loss functions listed previously, combined with the main task loss function for sentiment recognition (computing the Mean Absolute Error between \(y_{\text{final}}\) and the labels), we define our overall optimization objective as:
\begin{equation}
\begin{aligned}
\mathcal{L}_{\text{total}}
&= \mathcal{L}_{\text{task}}+\lambda_2\mathcal{L}_\text{cons}+\lambda_3\mathcal{L}_\text{conf}
\end{aligned}
\end{equation}
Hyperparameters $\lambda_2$ and $\lambda_3$ balance the strengths of the consensus space loss and conflict space loss.
\section{Experiments}
\paragraph{Datasets.}
To illustrate the wide-ranging utility of DynFlux in handling MER, MIR, and MSD tasks, we assess its performance on four key benchmarks: CMU-MOSI~\cite{Zadeh2016MOSI}, CMU-MOSEI~\cite{Zadeh2018MOSEI}, MIntRec~\cite{Zhang2022MIntRec}, and MMSD~\cite{cai2019mmsd}. \textbf{CMU-MOSI} comprises 2,199 monologue video clips divided into 1,284 training, 229 validation, and 686 testing samples, incorporating acoustic data sampled at 12.5 Hz alongside visual features at 15 Hz, and annotated with sentiment scores spanning from -3 (highly negative) to 3 (highly positive). Building on this, \textbf{CMU-MOSEI} offers a larger collection of 22,856 YouTube review excerpts, allocated as 16,326 for training, 1,871 for validation, and 4,659 for testing, with acoustic and visual modalities sampled at 20 Hz and 15 Hz respectively, while adhering to the identical sentiment annotation range. Complementing these, \textbf{MIntRec} provides 2,224 labeled video instances split into 1,334 training, 445 validation, and 445 testing sets, focusing on 11 intent categories tied to emotions or attitudes plus 9 linked to goal fulfillment. Rounding out the evaluation, \textbf{MMSD} draws from 24,635 Twitter-derived text-image pairs, segmented into 19,816 training, 2,410 validation, and 2,409 testing examples, featuring binary labels for sarcasm detection that highlight discrepancies between textual and visual elements in static multimodal pairs.
\paragraph{Evaluation Metric.}
For the CMU-MOSI and CMU-MOSEI datasets, we follow the evaluation approach established in prior works~\cite{li2023dmd,Hazarika2020MISA}, employing 7-class Accuracy (ACC-7), Binary Accuracy (ACC-2), F1-score (F1), and Mean Absolute Error (MAE) to assess sentiment analysis performance. On the MIntRec dataset, adhering to the standard protocol from studies~\cite{Zhang2022MIntRec,Sun2024CAGC}, we utilize Accuracy (ACC), F1-score (F1), Precision (P), and Recall (R) for intent recognition evaluation. For the MMSD dataset, following established practices in the field~\cite{cai2019mmsd,Wei2024G2SAM}, we adopt Accuracy (Acc), Precision (P), Recall (R), and F1-score (F1) as the evaluation metrics for sarcasm detection tasks.
\paragraph{Implementation details.}
For the CMU-MOSI and CMU-MOSEI datasets, we obtain 768-dimensional hidden states from BERT-base-uncased ~\cite{Devlin2019BERT} for textual representation. Visual features are represented by 35-dimensional facial action units obtained through Facet~\cite{Baltrusaitis2016OpenFace}, while acoustic features are processed into 74-dimensional vectors via COVAREP~\cite{Degottex2014COVAREP}. In the MIntRec dataset, the dimensionalities of text, visual, and acoustic features are set to 768, 256, and 768, respectively. For MMSD, we extract both textual and visual features using BERT-base and ViT, with all feature dimensions unified to 768. The optimal values for hyperparameters $\lambda_1$, $\lambda_2$, and $\lambda_3$ are all set to 0.1, the optimal Top-K is set to 5, and the optimal $\tau$ is set to 0.7. All experiments are implemented in PyTorch on an RTX 4090 GPU with 24GB memory, using a batch size of 16 and trained for 65 epochs.
\begin{table}[htbp]
  \centering
  \small                               
  \setlength{\tabcolsep}{3pt}        
  \renewcommand{\arraystretch}{1.06}   

  \begin{tabular}{l|c|cccc}
    \toprule
    \textbf{Methods} & \textbf{Setting} & $\mathbf{ACC_{7}}\!\!\uparrow$ & $\mathbf{ACC_{2}}\!\!\uparrow$ & \textbf{F1}$\!\uparrow$ & \textbf{MAE}$\!\downarrow$ \\
    \midrule
    EF--LSTM~\cite{Williams2018RecEmoVideo}      & \multirow{14}{*}{Aligned} & 47.4 & 78.2 & 77.9 & 0.620 \\
    LF--DNN~\cite{Williams2018RecEmoVideo}      &                           & 51.7 & 83.5 & 83.1 & 0.568 \\
    TFN~\cite{Zadeh2017TFN}          &                           & 50.9 & 80.4 & 80.7 & 0.574 \\
    LMF~\cite{liu2018lmf}            &                           & 52.3 & 84.7 & 84.5 & 0.564 \\
    MFN~\cite{zadeh2018mfn}          &                           & 50.8 & 84.0 & 84.0 & 0.574 \\
    Graph--MFN~\cite{Zadeh2018MOSEI}    &                           & 51.6 & 84.6 & 84.5 & 0.553 \\
    MFM~\cite{tsai2019lfmr}            &                           & 49.4 & 83.5 & 83.4 & 0.590 \\
    MulT~\cite{Tsai2019MulT}          &                           & 52.3 & 82.7 & 82.8 & 0.572 \\
    MISA\textsuperscript{*}~\cite{Hazarika2020MISA}   &         & 52.3 & 85.3 & 85.1 & 0.543 \\
    Self--MM\textsuperscript{*}~\cite{Yu2021SelfMM}     &         & 53.2 & 84.5 & 84.3 & 0.540 \\
    MMIN\textsuperscript{*}~\cite{Han2021MMIM}        &         & 50.1 & 83.6 & 83.5 & 0.580 \\
    GSCon\textsuperscript{*}~\cite{Shi2025GSCon}        &         & 51.2 & 83.4 & 83.3 & 0.544 \\
    DMD\textsuperscript{*}~\cite{li2023dmd}         &         & 52.4 & 84.8 & 84.7 & 0.546 \\
    EMOE\textsuperscript{*}~\cite{Fang2025EMOE}                         &         & 54.1 & 85.3 & 85.3 & 0.536 \\
    \rowcolor{LightCyan}
    \textbf{DynFlux(Ours)\textsuperscript{*}}              &         & \textbf{55.7} & \textbf{86.5} & \textbf{86.5} & \textbf{0.523} \\
    \midrule
    EF--LSTM~\cite{Williams2018RecEmoVideo}      & \multirow{14}{*}{Unaligned} & 46.3 & 76.1 & 75.9 & 0.594 \\
    LF--DNN~\cite{Williams2018RecEmoVideo}      &                             & 52.3 & 83.7 & 83.2 & 0.561 \\
    TFN~\cite{Zadeh2017TFN}          &                             & 50.2 & 84.2 & 84.0 & 0.573 \\
    LMF~\cite{liu2018lmf}            &                             & 51.9 & 83.8 & 83.9 & 0.565 \\
    MFN~\cite{zadeh2018mfn}          &                             & 51.3 & 83.2 & 83.3 & 0.567 \\
    Graph--MFN~\cite{Zadeh2018MOSEI}    &                             & 51.8 & 84.2 & 84.2 & 0.568 \\
    MFM~\cite{tsai2019lfmr}            &                             & 52.0 & 82.3 & 82.5 & 0.572 \\
    MulT~\cite{Tsai2019MulT}          &                             & 53.2 & 84.0 & 84.0 & 0.556 \\
    MISA\textsuperscript{*}~\cite{Hazarika2020MISA}   &           & 51.0 & 84.8 & 84.8 & 0.557 \\
    Self--MM\textsuperscript{*}~\cite{Yu2021SelfMM}     &           & 52.9 & 85.3 & 84.8 & 0.535 \\
    MMIN\textsuperscript{*}~\cite{Han2021MMIM}        &           & 52.6 & 81.5 & 81.3 & 0.578 \\
    GSCon\textsuperscript{*}~\cite{Shi2025GSCon}        &         & 53.5 & 84.3 & 84.2 & 0.524 \\
    DMD\textsuperscript{*}~\cite{li2023dmd}         &           & 53.1 & 84.7 & 84.7 & 0.536 \\
    EMOE\textsuperscript{*}~\cite{Fang2025EMOE}                          &          & 53.9 & 85.5 & 85.5 & 0.530 \\
    \rowcolor{LightCyan}
    \textbf{DynFlux(Ours)\textsuperscript{*}}                &          & \textbf{54.4} & \textbf{86.7} & \textbf{86.6} & \textbf{0.519} \\
    \bottomrule
  \end{tabular}
  \caption{Comparison on \textbf{CMU-MOSEI} dataset. Bold denotes the best result in each setting. ACC$_7$, ACC$_2$ and F1 are percentages. \textsuperscript{*}~indicates models that use BERT-based language features.}
  \label{tab:mosei}
\end{table}
\begin{table}[htbp]
\label{tab:mosi}
  \centering
  \small                    
  \setlength{\tabcolsep}{3pt}    
  \renewcommand{\arraystretch}{1.06}

  \begin{tabular}{l|c|cccc}
    \toprule
    \textbf{Methods} & \textbf{Setting} & $\mathbf{ACC_{7}}\!\!\uparrow$ & $\mathbf{ACC_{2}}\!\!\uparrow$ & \textbf{F1}$\!\uparrow$ & \textbf{MAE}$\!\downarrow$ \\
    \midrule
    EF--LSTM~\cite{Williams2018RecEmoVideo}      & \multirow{14}{*}{Aligned} & 33.7 & 75.3 & 75.2 & 1.386 \\
    LF--DNN~\cite{Williams2018RecEmoVideo}      &                           & 31.5 & 78.4 & 78.3 & 0.972 \\
    TFN~\cite{Zadeh2017TFN}          &                           & 31.9 & 78.8 & 78.9 & 0.953 \\
    LMF~\cite{liu2018lmf}            &                           & 36.9 & 78.7 & 78.7 & 0.931 \\
    MFN~\cite{zadeh2018mfn}          &                           & 35.6 & 78.4 & 78.4 & 0.964 \\
    Graph--MFN~\cite{Zadeh2018MOSEI}    &                           & 31.5 & 78.1 & 78.1 & 0.970 \\
    MCTN~\cite{Pham2019FoundInTranslation}          &                           & 33.1 & 79.3 & 80.0 & 0.963 \\
    MulT~\cite{Tsai2019MulT}          &                           & 35.1 & 80.0 & 80.1 & 0.936 \\
    MISA\textsuperscript{*}~\cite{Hazarika2020MISA}   &         & 41.8 & 84.2 & 84.2 & 0.754 \\
    Self--MM\textsuperscript{*}~\cite{Yu2021SelfMM}     &         & 45.3 & 84.9 & 84.9 & 0.738 \\
    MMIM\textsuperscript{*}~\cite{Han2021MMIM}        &         & 45.8 & 84.6 & 84.5 & 0.717 \\
    GSCon\textsuperscript{*}~\cite{Shi2025GSCon}        &         & 46.1 & 85.6 & 85.5 & 0.737 \\
    DMD\textsuperscript{*}~\cite{li2023dmd}         &         & 46.2 & 83.2 & 83.2 & 0.721 \\
    EMOE\textsuperscript{*}~\cite{Fang2025EMOE}                          &         & 47.7 & 85.4 & 85.4 & 0.710 \\
    \rowcolor{LightCyan}
    \textbf{DynFlux(Ours)\textsuperscript{*}}               &         & \textbf{48.5} & \textbf{86.9} & \textbf{86.9} & \textbf{0.695} \\
    \midrule
    EF--LSTM~\cite{Williams2018RecEmoVideo}      & \multirow{14}{*}{Unaligned} & 31.0 & 73.6 & 74.5 & 1.420 \\
    LF--DNN~\cite{Williams2018RecEmoVideo}      &                             & 32.5 & 78.2 & 78.3 & 0.987 \\
    TFN~\cite{Zadeh2017TFN}          &                             & 35.3 & 76.5 & 76.6 & 0.995 \\
    LMF~\cite{liu2018lmf}            &                             & 31.1 & 79.1 & 79.1 & 0.963 \\
    MFN~\cite{zadeh2018mfn}          &                             & 34.7 & 80.0 & 80.1 & 0.971 \\
    Graph--MFN~\cite{Zadeh2018MOSEI}    &                             & 34.4 & 79.4 & 79.2 & 0.930 \\
    MCTN~\cite{Pham2019FoundInTranslation}          &                             & 31.9 & 77.1 & 77.3 & 1.033 \\
    MulT~\cite{Tsai2019MulT}          &                             & 33.2 & 80.3 & 80.3 & 0.933 \\
    MISA\textsuperscript{*}~\cite{Hazarika2020MISA}   &           & 43.6 & 83.8 & 83.9 & 0.742 \\
    Self--MM\textsuperscript{*}~\cite{Yu2021SelfMM}     &           & 45.7 & 83.4 & 83.6 & 0.724 \\
    MMIM\textsuperscript{*}~\cite{Han2021MMIM}        &           & 45.9 & 83.4 & 83.4 & 0.777 \\
    GSCon\textsuperscript{*}~\cite{Shi2025GSCon}        &         & 46.5 & 86.1 & 86.0 & 0.731 \\
    DMD\textsuperscript{*}~\cite{li2023dmd}         &           & 46.7 & 84.0 & 84.0 & 0.721 \\
    EMOE\textsuperscript{*}~\cite{Fang2025EMOE}                          &          & 47.8 & 85.4 & 85.3 & 0.697 \\
    \rowcolor{LightCyan}
    \textbf{DynFlux(Ours)\textsuperscript{*}}                &          & \textbf{48.6} & \textbf{87.0} & \textbf{86.9} & \textbf{0.689} \\
    \bottomrule
  \end{tabular}
  \caption{Comparison on \textbf{CMU-MOSI} dataset. Bold indicates the best. ACC$_7$, ACC$_2$ and F1 are percentages. *~denotes models with BERT-based language features.}
  \label{tab:mosi}
\end{table}
\subsection{Experimental Results and Analysis}
\paragraph{Results on the MER datasets.}
We evaluate DynFlux on the CMU-MOSEI and CMU-MOSI benchmarks under aligned and unaligned settings, comparing it with state-of-the-art MER methods such as TFN~\cite{Zadeh2017TFN}, MulT~\cite{Tsai2019MulT}, MISA~\cite{Hazarika2020MISA}, Self MM~\cite{Yu2021SelfMM}, GSCon~\cite{Shi2025GSCon}, DMD~\cite{li2023dmd}, and EMOE~\cite{Fang2025EMOE}. As shown in Tab.~\ref{tab:mosei} and Tab.~\ref{tab:mosi}, DynFlux outperforms all methods across all metrics. On CMU-MOSEI, it achieves 55.7\% ACC$_7$, 86.5\% ACC$_2$, and 86.5\% F1 with a MAE of 0.523, surpassing EMOE by 1.6 points on ACC$_7$, 1.2 points on ACC$_2$, and 0.8 points on F1, while reducing MAE by 0.013. In the unaligned setting, it improves ACC$_2$ to 86.7\% and F1 to 86.6\%, with MAE reduced to 0.519. On CMU-MOSI, DynFlux sets new state-of-the-art results, achieving 48.5\% ACC$_7$, 86.9\% ACC$_2$, and 86.9\% F1 with 0.695 MAE in the aligned setting, and 48.6\% ACC$_7$ and 87.0\% ACC$_2$ with 0.689 MAE in the unaligned setting.

The advantages of DynFlux are clear from these results. By explicitly modeling cross-modal consensus and conflict, and routing them to the fusion network, DynFlux amplifies emotion cues supported by multiple modalities, using consensus to establish a stable emotional foundation and conflict to refine predictions. Unlike methods such as EMOE that rely on a fixed fusion scheme, DynFlux adapts its fusion process to both input and time, which is especially effective for handling frequent modality shifts in long CMU-MOSEI videos. Its temporal modeling tracks how consensus and conflict evolve across frames, guiding the model to focus on short but key emotional segments while avoiding noise. Furthermore, its hierarchical guidance mechanism propagates consensus and conflict information throughout the pipeline. At the frame and sequence levels, it recalibrates modalities to ensure each one contributes reliably, while at the cross-modal fusion level, it adjusts modality weights based on global patterns. This multi-level, consensus and conflict-aware design allows DynFlux to better exploit multimodal interactions and achieve more stable emotion recognition.

\paragraph{Results on the MIR datasets.}
\begin{table}[htbp]
  \centering
  \small
  \setlength{\tabcolsep}{6pt}
  \renewcommand{\arraystretch}{1.06}

  \begin{tabularx}{\linewidth}{l|XXXX}
    \toprule
    \textbf{Methods} & \textbf{ACC}$\uparrow$ & \textbf{F1}$\uparrow$ & \textbf{P}$\uparrow$ & \textbf{R}$\uparrow$ \\
    \midrule
    MAG-BERT~\cite{Rahman2020MAG} & 70.34 & 68.19 & 68.31 & 69.36 \\
    MulT~\cite{Tsai2019MulT}      & 72.58 & 69.36 & 70.73 & 69.47 \\
    MISA~\cite{Hazarika2020MISA}  & 72.36 & 70.57 & 71.24 & 70.41 \\
    GSCon~\cite{Shi2025GSCon}     & 72.43 & 70.62 & 71.54 & 70.61 \\
    EMOE~\cite{Fang2025EMOE}      & 72.58 & 70.73 & 72.08 & 70.86 \\
    \rowcolor{LightCyan}
    \textbf{DynFlux (Ours)}        & \textbf{73.01} & \textbf{71.32} & \textbf{73.56} & \textbf{71.44} \\
    \bottomrule
  \end{tabularx}

  \caption{Comparison on the \textbf{MIntRec} dataset. Bold denotes the best result in each column. ACC, F1, P and R values are shown as percentages.}
  \label{tab:mintrec}
\end{table}

As shown in Tab.~\ref{tab:mintrec}, DynFlux also achieves the best performance on the MIntRec dataset. Our model reaches 73.01\% ACC and 71.32\% F1, outperforming the strong EMOE baseline by 0.43 and 0.59 points respectively, and obtains clear gains on both precision (73.56\%) and recall (71.44\%). These results indicate that the consensus and conflict-aware design of DynFlux not only benefits multimodal emotion recognition, but also transfers well to intent understanding scenarios, where it helps the model exploit complementary textual, visual and acoustic cues for more accurate predictions. This is because intent recognition is highly correlated with emotion recognition, and MIR requires not only simple emotion classification but also a deeper understanding of emotions. By performing more insightful modeling of multimodal information, DynFlux achieves superior performance and robustness.

\paragraph{Results on the MSD datasets.}
\begin{table}[htbp]
  \centering
  \small
  \setlength{\tabcolsep}{6pt}
  \renewcommand{\arraystretch}{1.06}

  \begin{tabularx}{\linewidth}{l|XXXX}
    \toprule
    \textbf{Methods} & \textbf{ACC}$\uparrow$ & \textbf{F1}$\uparrow$ & \textbf{P}$\uparrow$ & \textbf{R}$\uparrow$ \\
    \midrule
    DMD~\cite{li2023dmd}          & 86.45 & 84.23 & 82.71 & 85.12 \\
    GSCon~\cite{Shi2025GSCon}     & 87.12 & 84.96 & 83.47 & 86.08 \\
    EMOE~\cite{Fang2025EMOE}      & 87.63 & 85.42 & 83.92 & 86.71 \\
    \rowcolor{LightCyan}
    \textbf{DynFlux (Ours)}        & \textbf{88.32} & \textbf{86.21} & \textbf{84.63} & \textbf{87.53} \\
    \bottomrule
  \end{tabularx}

  \caption{Comparison on the \textbf{MMSD} dataset. Bold denotes the best result in each column. ACC, F1, P and R values are shown as percentages.}
  \label{tab:mmsd}
\end{table}

As shown in Tab.~\ref{tab:mmsd}, DynFlux also achieves the best performance on the MMSD dataset. It attains 88.32\% ACC and 86.21\% F1, surpassing the strong EMOE baseline by 0.69 and 0.79 points respectively, and further improves precision and recall to 84.63\% and 87.53\%. Compared with DMD and GSCon, the margins are even larger, which suggests that the consensus and conflict-aware design of DynFlux is effective not only on emotion and intent benchmarks, but also transfers well to multimodal sentiment detection in the more diverse scenarios covered by MMSD. Indeed, the sarcasm detection task relies more heavily on conflicting information. This specific type of sentiment analysis requires more complex multimodal information interaction. DynFlux effectively addresses this by explicitly modeling the extraction and interaction of both consensus and conflict within multimodal data.

\subsection{Ablation study}
\begin{table}[htbp]
  \centering
  \small
  \setlength{\tabcolsep}{4pt}
  \renewcommand{\arraystretch}{1.06}

  \begin{tabular}{cccc|cccc}
    \toprule
    \textbf{Graph} & \textbf{Cons.} & \textbf{Conf.} & \textbf{DRF} 
    & \textbf{ACC$_7$}$\!\uparrow$ & \textbf{ACC$_2$}$\!\uparrow$ & \textbf{F1}$\!\uparrow$ & \textbf{MAE}$\!\downarrow$ \\
    \midrule
      &   &   &   & 52.0 & 84.3 & 84.1 & 0.548 \\  
    \checkmark & \checkmark &        & \checkmark & 53.6 & 85.1 & 85.0 & 0.538 \\  
    \checkmark &            & \checkmark & \checkmark & 53.1 & 84.9 & 84.7 & 0.541 \\  
    \checkmark & \checkmark & \checkmark &           & 54.3 & 85.5 & 85.6 & 0.534 \\  
    \rowcolor{LightCyan}
    \checkmark & \checkmark & \checkmark & \checkmark 
               & \textbf{55.7} & \textbf{86.5} & \textbf{86.5} & \textbf{0.523} \\  
    \bottomrule
  \end{tabular}

  \caption{Ablation study of the key components in DynFlux on the \textbf{CMU--MOSEI} dataset (aligned setting). 
  "Graph" denotes the consensus-conflict temporal graph branch.}
  \label{tab:ablation_mosei}
\end{table}

\begin{table}[htbp]
  \centering
  \small
  \setlength{\tabcolsep}{4pt}
  \renewcommand{\arraystretch}{1.06}

  \begin{tabular}{cccc|cccc}
    \toprule
    \textbf{SLM} & \textbf{FLM} & \textbf{CMM} & \textbf{MMD} 
    & \textbf{ACC$_7$}$\!\uparrow$ & \textbf{ACC$_2$}$\!\uparrow$ & \textbf{F1}$\!\uparrow$ & \textbf{MAE}$\!\downarrow$ \\
    \midrule
                 &             &             &           
                 & 53.8 & 85.1 & 85.0 & 0.537 \\  
    \checkmark   &             & \checkmark   & \checkmark
                 & 55.0 & 86.0 & 86.0 & 0.528 \\  
                 & \checkmark   & \checkmark   & \checkmark
                 & 54.9 & 85.9 & 85.8 & 0.529 \\  
    \checkmark   & \checkmark   &             & \checkmark
                 & 55.1 & 86.1 & 86.0 & 0.527 \\  
    \checkmark   & \checkmark   & \checkmark   &           
                 & 54.9 & 85.9 & 85.9 & 0.529 \\  
    \rowcolor{LightCyan}
    \checkmark   & \checkmark   & \checkmark   & \checkmark
                 & \textbf{55.7} & \textbf{86.5} & \textbf{86.5} & \textbf{0.523} \\  
    \bottomrule
  \end{tabular}

  \caption{Ablation study of the three-level modulation (SLM, FLM, CMM) and the MMD-based constraints on the \textbf{CMU--MOSEI~\cite{Zadeh2018MOSEI}} dataset (aligned setting).}
  \label{tab:ablation_mod_mmd}
\end{table}

\begin{figure*}[t] 
  \centering
  \begin{subfigure}{0.19\textwidth}
    \centering
    \includegraphics[width=\textwidth]{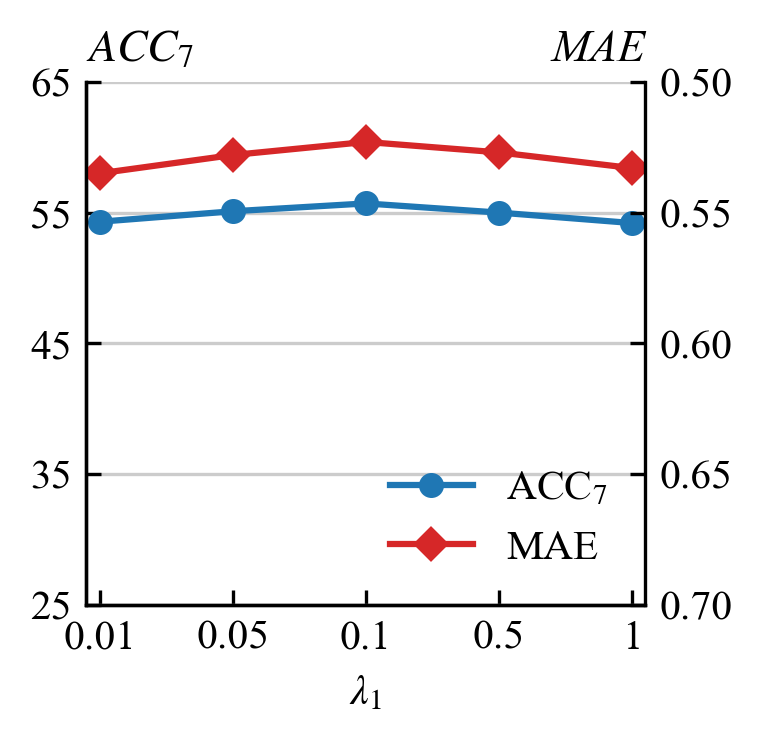}
    \caption{$\lambda_1$}
    \label{fig:sens_lambda1}
  \end{subfigure}\hfill
  \begin{subfigure}{0.19\textwidth}
    \centering
    \includegraphics[width=\textwidth]{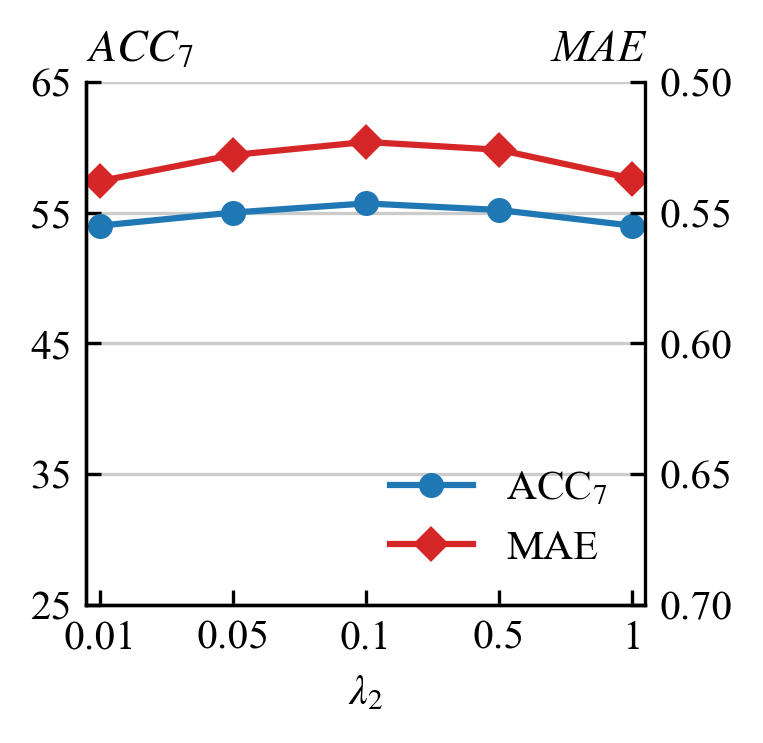}
    \caption{$\lambda_2$}
    \label{fig:sens_lambda2}
  \end{subfigure}\hfill
  \begin{subfigure}{0.19\textwidth}
    \centering
    \includegraphics[width=\textwidth]{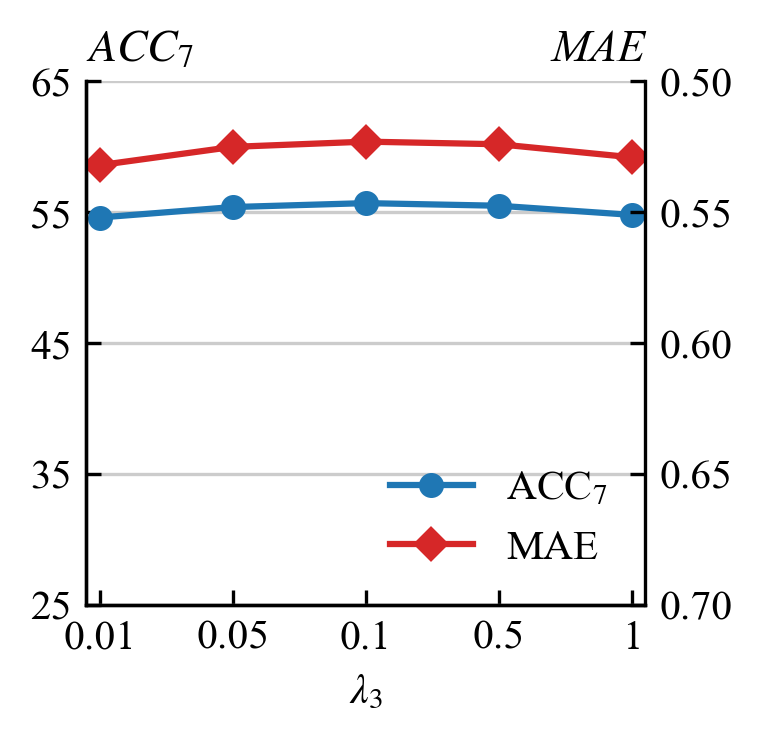}
    \caption{$\lambda_3$}
    \label{fig:sens_lambda3}
  \end{subfigure}\hfill
  \begin{subfigure}{0.19\textwidth}
    \centering
    \includegraphics[width=\textwidth]{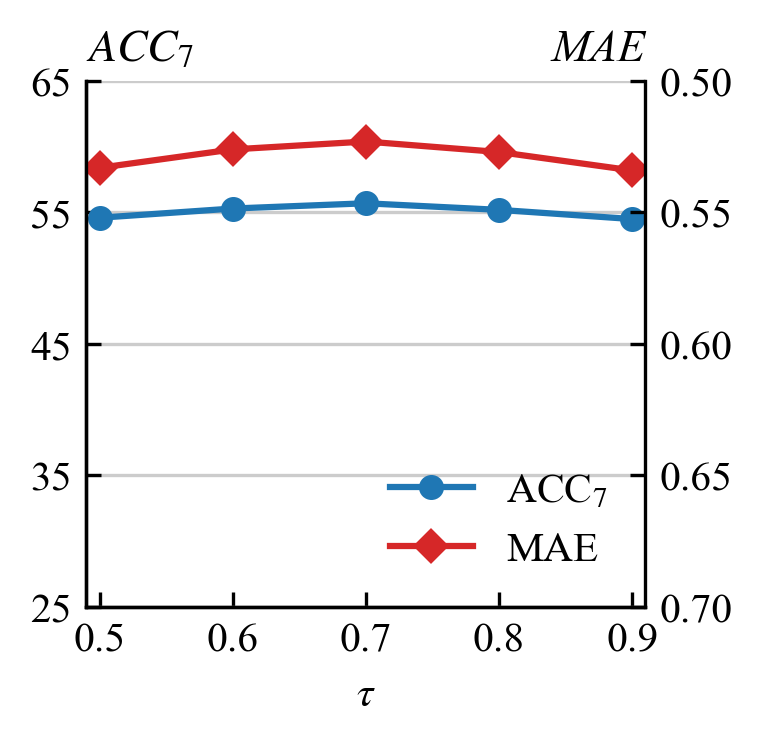}
    \caption{$\tau$}
    \label{fig:sens_tau}
  \end{subfigure}\hfill
  \begin{subfigure}{0.19\textwidth}
    \centering
    \includegraphics[width=\textwidth]{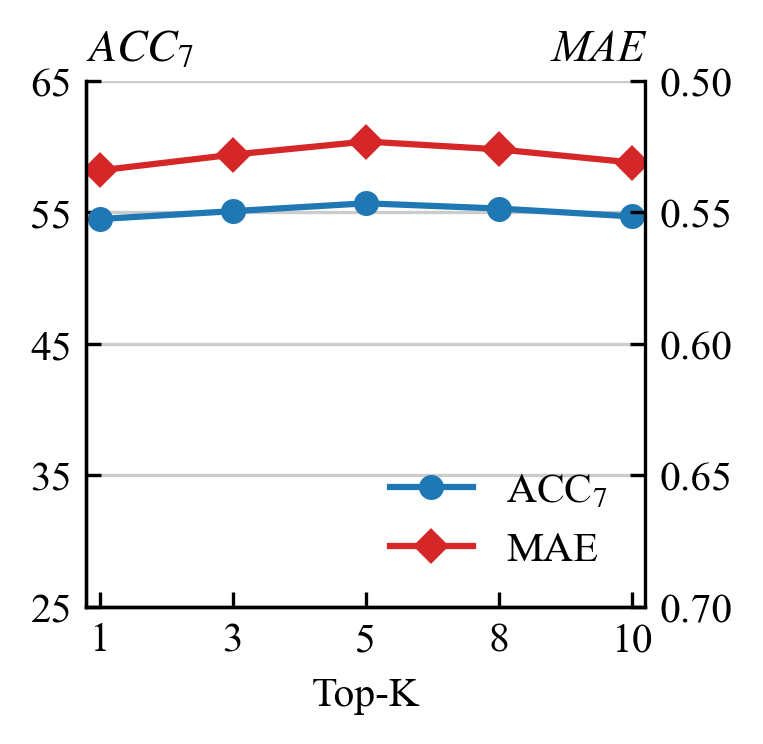}
    \caption{Top-K}
    \label{fig:sens_topk}
  \end{subfigure}
  
  \caption{Sensitivity of DynFlux to different hyper-parameters: (a) frame-level modulation weight $\lambda_1$, (b) consensus-space loss weight $\lambda_2$, (c) conflict-space loss weight $\lambda_3$, (d) similarity threshold $\tau$, and (e) number of nearest neighbors Top-K.}
  \label{fig:sensitivity_analysis}
\end{figure*}
We evaluate the effect of the key components in DynFlux on CMU--MOSEI, including the consensus--conflict temporal graph branch, the consensus and conflict subgraphs, and the DRF module, as summarized in Tab.~\ref{tab:ablation_mosei}. 
\textbf{Firstly}, introducing the graph branch brings clear performance gains over the backbone without graphs: ACC$_7$ increases from 52.0\% to 55.7\%, ACC$_2$ and F1 improve by 2.2 and 2.4 points, and MAE is reduced from 0.548 to 0.523. 
This confirms that explicitly modeling cross-modal relations provides complementary cues to the standard MER pipeline. 
\textbf{Secondly}, using only the consensus or only the conflict subgraph is consistently inferior to the full graph, where ACC$_7$ remains below 54.0\% and MAE stays above 0.538. 
The full configuration that jointly leverages consensus and conflict therefore captures more comprehensive multimodal interactions. 
\textbf{Thirdly}, removing DRF and simply fusing the two streams leads to noticeable degradation (ACC$_7$: 55.7\% $\rightarrow$ 54.3\%, MAE: 0.523 $\rightarrow$ 0.534), indicating that the content-aware routing between consensus and conflict representations is crucial for exploiting their complementary strengths and suppressing misleading signals.

We also evaluate the effect of the three-level modulation and the MMD-based constraints on CMU--MOSEI, as summarized in Tab.~\ref{tab:ablation_mod_mmd}. \textbf{Firstly}, adding hierarchical modulation and MMD on top of the backbone clearly improves performance: compared with the variant without any modulation or MMD, ACC$_7$ increases from 53.8\% to 55.7\%, ACC$_2$ and F1 rise from 85.1\% and 85.0\% to 86.5\%, and MAE is reduced from 0.537 to 0.523, showing that feeding consensus--conflict signals back into the MER pipeline is beneficial. \textbf{Secondly}, disabling any single modulation level (SLM, FLM or CMM) consistently degrades the results relative to the full model: ACC$_7$ falls to 55.0\% without SLM, 54.9\% without FLM and 55.1\% without CMM, while MAE rises to around 0.527--0.529, indicating that these three levels provide complementary gains, with SLM having the most visible impact on F1 and MAE. \textbf{Thirdly}, removing the MMD constraints while keeping SLM, FLM and CMM also leads to noticeable performance drops (ACC$_7$: 55.7\% $\rightarrow$ 54.9\%, MAE: 0.523 $\rightarrow$ 0.529), which suggests that regularizing the consensus and conflict subspaces helps learn more discriminative guidance signals for DynFlux.

\subsection{Sensitivity Analysis}
\newcommand{\sensfigwidth}{0.2\linewidth}
To demonstrate the robustness of the proposed DynFlux model, we conduct
a sensitivity analysis over the loss weights $\lambda_1$, $\lambda_2$,
and $\lambda_3$, as well as the graph construction hyper-parameters
Top-K and $\tau$. For each setting, we vary one hyper-parameter while
keeping all the others fixed to the values adopted in the main
experiments, and report the corresponding ACC$_7$ and MAE. The results in
Figs.~\ref{fig:sens_lambda1}--\ref{fig:sens_topk}
show that the overall performance curves remain relatively stable,
indicating that DynFlux is not overly sensitive to these
hyper-parameters.
\paragraph{Sensitivity to $\lambda_1$.}
Fig.~\ref{fig:sens_lambda1} shows the impact of the frame-level
modulation weight $\lambda_1$.
Both ACC$_7$ and MAE vary only slightly across a wide range of values.
The best performance is obtained around $\lambda_1\in[0.1,0.5]$, while
very small ($0.01$) or very large ($1.0$) weights lead to a mild
degradation. This suggests that the proposed modulation term provides
consistent benefits as long as its contribution is kept within a
reasonable range.
\paragraph{Sensitivity to $\lambda_2$.}
As illustrated in Fig.~\ref{fig:sens_lambda2}, the consensus-space loss
weight $\lambda_2$ exhibits a similar trend.
Increasing $\lambda_2$ from $0.01$ to $0.1$ slightly improves
ACC$_7$ and reduces MAE, after which further enlarging the weight does
not bring additional gains and may even lead to a small drop.
This confirms that a moderate consensus constraint is sufficient to
regularize the model without over-smoothing the representations.
\paragraph{Sensitivity to $\lambda_3$.}
Fig.~\ref{fig:sens_lambda3} reports the sensitivity with respect to the
conflict-space loss weight $\lambda_3$.
The model again achieves its best performance near $\lambda_3=0.1$,
with both metrics remaining stable when $\lambda_3$ varies within an
order of magnitude. Extreme values slightly deteriorate the results,
indicating that excessively strong or weak conflict modeling is
undesirable, whereas a balanced setting effectively complements the
consensus term.
\paragraph{Sensitivity to $\tau$.}
The effect of the similarity threshold $\tau$ for edge pruning is shown
in Fig.~\ref{fig:sens_tau}.
When $\tau$ increases from $0.5$ to $0.7$, ACC$_7$ improves and MAE
decreases, suggesting that removing very weak cross-modal connections
helps denoise the graph. Beyond $\tau=0.7$, performance becomes slightly
worse, which implies that overly aggressive pruning may discard
informative relations. Overall, the curves are flat, demonstrating that
DynFlux is robust to the choice of $\tau$ around the default value.
\paragraph{Sensitivity to Top-K.}
Finally, Fig.~\ref{fig:sens_topk} presents the sensitivity against the
number of retained neighbors Top-K.
Both metrics gradually improve when increasing Top-K from $1$ to $5$,
showing the benefit of incorporating more cross-modal interactions.
Further enlarging Top-K to $8$ or $10$ does not provide additional
gains and even slightly harms performance, likely due to the inclusion
of noisy edges.
These results validate that a moderately sparse graph (Top-K $\approx 5$)
offers a good trade-off between information aggregation and robustness.

\section{Conclusion}
In this paper, we proposed DynFlux, which models cross-modal consensus and conflict with temporal heterogeneous graphs and adaptively fuses modalities via disentangled routing. By explicitly capturing both agreement and discrepancy across modalities over time, DynFlux provides a more structured and interpretable way to characterize multimodal interactions. Experiments on CMU-MOSI, CMU-MOSEI, MIntRec, and MMSD show consistent gains over strong baselines, indicating that consensus–conflict structure is a useful inductive bias for emotion and intent understanding.

\bibliographystyle{ACM-Reference-Format}
\bibliography{sample-base}

@String{Computing = "Computing" }

@String{Computer = "{IEEE} Computer" }

@String{Springer = "Springer-Verlag" }

@inproceedings{Liang2021Attention,
  author    = {Tao Liang and Guosheng Lin and Lei Feng and Yan Zhang and Fengmao Lv},
  title     = {Attention Is Not Enough: Mitigating the Distribution Discrepancy in Asynchronous Multimodal Sequence Fusion},
  booktitle = ICCV,
  pages     = {8148--8156},
  year      = {2021}
}

@inproceedings{Lv2021PMR,
  author    = {Fengmao Lv and Xiang Chen and Yanyong Huang and Lixin Duan and Guosheng Lin},
  title     = {Progressive Modality Reinforcement for Human Multimodal Emotion Recognition From Unaligned Multimodal Sequences},
  booktitle = CVPR,
  pages     = {2554--2562},
  year      = {2021},
  doi       = {10.1109/CVPR46437.2021.00258}
}

@article{Zadeh2016MOSI,
  author  = {Amir Zadeh and Rowan Zellers and Eli Pincus and Louis-Philippe Morency},
  title   = {Multimodal Sentiment Intensity Analysis in Videos: Facial Gestures and Verbal Messages},
  journal = {IEEE Intelligent Systems},
  volume  = {31},
  number  = {6},
  pages   = {82--88},
  year    = {2016},
  doi     = {10.1109/MIS.2016.94}
}

@article{Poria2023Iceberg,
  author  = {Soujanya Poria and Devamanyu Hazarika and Navonil Majumder and Rada Mihalcea},
  title   = {Beneath the Tip of the Iceberg: Current Challenges and New Directions in Sentiment Analysis Research},
  journal = {IEEE Transactions on Affective Computing},
  volume  = {14},
  number  = {1},
  pages   = {108--132},
  year    = {2023},
  doi     = {10.1109/TAFFC.2020.3038167}
}

@article{Petrovica2017ATS,
  author  = {Sintija Petrovica and Alla Anohina-Naumeca and Hazim Kemal Ekenel},
  title   = {Emotion Recognition in Affective Tutoring Systems: Collection of Ground-Truth Data},
  journal = {Procedia Computer Science},
  volume  = {104},
  pages   = {437--444},
  year    = {2017},
  doi     = {10.1016/j.procs.2017.01.157}
}

@inproceedings{Melville2009Blogs,
  author    = {Prem Melville and Wojciech Gryc and Richard D. Lawrence},
  title     = {Sentiment Analysis of Blogs by Combining Lexical Knowledge with Text Classification},
  booktitle = {Proceedings of the 15th ACM SIGKDD International Conference on Knowledge Discovery and Data Mining (KDD '09)},
  pages     = {1275--1284},
  year      = {2009},
  address   = {Paris, France},
  publisher = {ACM},
  doi       = {10.1145/1557019.1557156}
}

@article{Liu2017FEERHRI,
  author  = {Zhentao Liu and Min Wu and Weihua Cao and Luefeng Chen and Jianping Xu and Ri Zhang and Mengtian Zhou and Junwei Mao},
  title   = {A Facial Expression Emotion Recognition Based Human-robot Interaction System},
  journal = {IEEE/CAA Journal of Automatica Sinica},
  volume  = {4},
  number  = {4},
  pages   = {668--676},
  year    = {2017},
  doi     = {10.1109/JAS.2017.7510622}
}

@inproceedings{Hazarika2020MISA,
  author    = {Devamanyu Hazarika and Roger Zimmermann and Soujanya Poria},
  title     = {MISA: Modality-Invariant and -Specific Representations for Multimodal Sentiment Analysis},
  booktitle = ACMMM,
  pages     = {1122--1131},
  year      = {2020},
  doi       = {10.1145/3394171.3413678}
}

@inproceedings{Tsai2019MulT,
  author    = {Yao{-}Hung Hubert Tsai and Shaojie Bai and Paul Pu Liang and J. Zico Kolter and Louis{-}Philippe Morency and Ruslan Salakhutdinov},
  title     = {Multimodal Transformer for Unaligned Multimodal Language Sequences},
  booktitle = {Proceedings of the 57th Annual Meeting of the Association for Computational Linguistics},
  pages     = {6558--6569},
  year      = {2019},
  address   = {Florence, Italy},
  publisher = {Association for Computational Linguistics},
  doi       = {10.18653/v1/P19-1656}
}

@inproceedings{Zadeh2017TFN,
  author    = {Amir Zadeh and Minghai Chen and Soujanya Poria and Erik Cambria and Louis{-}Philippe Morency},
  title     = {Tensor Fusion Network for Multimodal Sentiment Analysis},
  booktitle = {Proceedings of the 2017 Conference on Empirical Methods in Natural Language Processing},
  pages     = {1103--1114},
  year      = {2017},
  address   = {Copenhagen, Denmark},
  publisher = {Association for Computational Linguistics},
  doi       = {10.18653/v1/D17-1115}
}

@article{Huan2024TriSAT,
  author  = {Ruohong Huan and Guowei Zhong and Peng Chen and Ronghua Liang},
  title   = {TriSAT: Trimodal Representation Learning for Multimodal Sentiment Analysis},
  journal = {IEEE/ACM Transactions on Audio, Speech, and Language Processing},
  volume  = {32},
  pages   = {4105--4120},
  year    = {2024},
  doi     = {10.1109/TASLP.2024.3458812}
}

@article{Shi2025GSCon,
  author  = {Qinghongya Shi and Mang Ye and Wenke Huang and Bo Du and Xiaofen Zong},
  title   = {Gradient and Structure Consistency in Multimodal Emotion Recognition},
  journal = {IEEE Transactions on Image Processing},
  volume  = {34},
  pages   = {6180--6191},
  year    = {2025},
  doi     = {10.1109/TIP.2025.3608664}
}

@article{Zhong2025CMC,
  author  = {Guowei Zhong and Junjie Li and Huaiyu Zhu and Ruohong Huan and Yun Pan},
  title   = {Calibrating Multimodal Consensus for Emotion Recognition},
  journal = {arXiv preprint arXiv:2510.20256},
  year    = {2025},
  url     = {https://arxiv.org/abs/2510.20256}
}

@inproceedings{Fang2025EMOE,
  author    = {Yiyang Fang and Wenke Huang and Guancheng Wan and Kehua Su and Mang Ye},
  title     = {EMOE: Modality-Specific Enhanced Dynamic Emotion Experts},
  booktitle = {Proceedings of the IEEE/CVF Conference on Computer Vision and Pattern Recognition (CVPR)},
  pages     = {14314--14324},
  year      = {2025},
  doi       = {10.1109/CVPR52734.2025.01335}
}

@inproceedings{Zadeh2018MOSEI,
  author    = {AmirAli Bagher Zadeh and Paul Pu Liang and Soujanya Poria and Erik Cambria and Louis-Philippe Morency},
  title     = {Multimodal Language Analysis in the Wild: {CMU\mbox{-}MOSEI} Dataset and Interpretable Dynamic Fusion Graph},
  booktitle = {Proceedings of the 56th Annual Meeting of the Association for Computational Linguistics (Volume 1: Long Papers)},
  pages     = {2236--2246},
  year      = {2018},
  address   = {Melbourne, Australia},
  publisher = {Association for Computational Linguistics},
  doi       = {10.18653/v1/P18-1208},
  url       = {https://aclanthology.org/P18-1208/}
}

@inproceedings{Zhang2022MIntRec,
  author    = {Hanlei Zhang and Hua Xu and Xin Wang and Qianrui Zhou and Shaojie Zhao and Jiayan Teng},
  title     = {MIntRec: A New Dataset for Multimodal Intent Recognition},
  booktitle = ACMMM,
  pages     = {1688--1697},
  year      = {2022},
  doi       = {10.1145/3503161.3547906}
}

@inproceedings{liu2018lmf,
  author    = {Zhun Liu and Ying Shen and Varun Bharadhwaj Lakshminarasimhan and Paul Pu Liang and AmirAli Bagher Zadeh and Louis-Philippe Morency},
  title     = {Efficient Low-rank Multimodal Fusion with Modality-Specific Factors},
  booktitle = ACL,
  pages     = {2247--2256},
  year      = {2018},
  doi       = {10.18653/v1/P18-1209}
}

@inproceedings{zadeh2018mfn,
  author    = {Amir Zadeh and Paul Pu Liang and Navonil Mazumder and Soujanya Poria and Erik Cambria and Louis-Philippe Morency},
  title     = {Memory Fusion Network for Multi-view Sequential Learning},
  booktitle = AAAI,
  pages     = {5634--5641},
  year      = {2018},
  doi       = {10.1609/aaai.v32i1.12021}
}

@inproceedings{morency2011towards,
  author    = {Louis{-}Philippe Morency and Rada Mihalcea and Payal Doshi},
  title     = {Towards Multimodal Sentiment Analysis: Harvesting Opinions From the Web},
  booktitle = ICMI,
  pages     = {169--176},
  year      = {2011},
  doi       = {10.1145/2070481.2070509}
}

@inproceedings{peng2022bml,
  author    = {Xiaokang Peng and Yake Wei and Andong Deng and Dong Wang and Di Hu},
  title     = {Balanced Multimodal Learning via On-the-Fly Gradient Modulation},
  booktitle = CVPR,
  pages     = {8238--8247},
  year      = {2022},
  doi       = {10.1109/CVPR52688.2022.00806}
}

@inproceedings{li2023dmd,
  author    = {Yong Li and Yuanzhi Wang and Zhen Cui},
  title     = {Decoupled Multimodal Distilling for Emotion Recognition},
  booktitle = CVPR,
  pages     = {6631--6640},
  year      = {2023}
}

@inproceedings{tsai2019lfmr,
  author    = {Yao{-}Hung Hubert Tsai and Paul Pu Liang and Amir Zadeh and Louis{-}Philippe Morency and Ruslan Salakhutdinov},
  title     = {Learning Factorized Multimodal Representations},
  booktitle = ICLR,
  year      = {2019},
  note      = {arXiv:1806.06176}
}

@inproceedings{yang2022drler,
  author    = {Dingkang Yang and Shuai Huang and Haopeng Kuang and Yangtao Du and Lihua Zhang},
  title     = {Disentangled Representation Learning for Multimodal Emotion Recognition},
  booktitle = ACMMM,
  pages     = {1642--1651},
  year      = {2022},
  doi       = {10.1145/3503161.3547754}
}

@inproceedings{wu2024nextgpt,
  author    = {Shengqiong Wu and Hao Fei and Leigang Qu and Wei Ji and Tat{-}Seng Chua},
  title     = {{NExT\mbox{-}GPT}: Any\mbox{-}to\mbox{-}Any Multimodal {LLM}},
  booktitle = ICML,
  pages     = {53366--53397},
  year      = {2024},
  url       = {https://proceedings.mlr.press/v235/wu24e.html}
}

@inproceedings{wang2024wisdom,
  author    = {Wenbin Wang and Liang Ding and Li Shen and Yong Luo and Han Hu and Dacheng Tao},
  title     = {WisdoM: Improving Multimodal Sentiment Analysis by Fusing Contextual World Knowledge},
  booktitle = ACMMM,
  pages     = {2282--2291},
  year      = {2024},
  doi       = {10.1145/3664647.3681403}
}

@article{shi2023c3n,
  title   = {Cross-modal Cognitive Consensus guided Audio-Visual Segmentation},
  author  = {Shi, Zhaofeng and Wu, Qingbo and Meng, Fanman and Xu, Linfeng and Li, Hongliang},
  journal = {arXiv preprint arXiv:2310.06259},
  year    = {2023}
}

@article{dai2024gccl,
  title   = {Multi-modal Graph Context Extraction and Consensus-aware Learning for Emotion Recognition in Conversation},
  author  = {Dai, Yijing and Li, Jinxing and Li, Yingjian and Lu, Guangming},
  journal = {Knowledge-Based Systems},
  volume  = {298},
  pages   = {111954},
  year    = {2024},
  publisher = {Elsevier}
}

@inproceedings{du2025stil,
  title     = {STiL: Semi-supervised Tabular-Image Learning for Comprehensive Task-Relevant Information Exploration in Multimodal Classification},
  author    = {Du, Siyi and Luo, Xinzhe and O'Regan, Declan P. and Qin, Chen},
  booktitle = {Proceedings of the IEEE/CVF Conference on Computer Vision and Pattern Recognition (CVPR)},
  year      = {2025}
}

@article{hao2025unix,
  title   = {Uni-X: Mitigating Modality Conflict with a Two-End-Separated Architecture for Unified Multimodal Models},
  author  = {Hao, Jitai and Liu, Hao and Xiao, Xinyan and Huang, Qiang and Yu, Jun},
  journal = {arXiv preprint arXiv:2509.24365},
  year    = {2025}
}

@article{liu2025codaf,
  title   = {Cross-modal Offset-guided Dynamic Alignment and Fusion for Weakly Aligned UAV Object Detection},
  author  = {Liu, Zongzhen and Luo, Hui and Wang, Zhixing and Wei, Yuxing and Zuo, Haorui and Zhang, Jianlin},
  journal = {arXiv preprint arXiv:2506.16737},
  year    = {2025}
}

@article{srikanth2025mixed,
  title   = {Mixed Signals: Understanding Model Disagreement in Multimodal Empathy Detection},
  author  = {Srikanth, Maya and Chen, Run and Hirschberg, Julia},
  journal = {arXiv preprint arXiv:2505.13979},
  year    = {2025}
}

@article{gao2025mcan,
  title   = {Multi-level Conflict-Aware Network for Multi-modal Sentiment Analysis},
  author  = {Gao, Yubo and Wu, Haotian and Zhang, Lei},
  journal = {arXiv preprint arXiv:2502.09675},
  year    = {2025}
}

@inproceedings{shi2025raven,
  title     = {RAVEN: Query-Guided Representation Alignment for Robust Vision--Language Question Answering},
  author    = {Shi, Zhaofeng and Wu, Qingbo and Meng, Fanman and Li, Hongliang},
  booktitle = {Proceedings of the 2025 Conference on Empirical Methods in Natural Language Processing (EMNLP)},
  year      = {2025},
  note      = {to appear},
  url       = {https://arxiv.org/abs/2504.12345}
}

@inproceedings{karri2025ugcemt,
  title     = {Uncertainty-Guided Cross Attention Ensemble Mean Teacher for Semi-supervised Medical Image Segmentation},
  author    = {Meghana Karri and Amit Soni Arya and Koushik Biswas and Nicolo Gennaro and Vedat Cicek and Gorkem Durak and Yuri S. Velichko and Ulas Bagci},
  booktitle = {Proceedings of the IEEE/CVF Winter Conference on Applications of Computer Vision (WACV)},
  year      = {2025},
  pages     = {to appear},
  url       = {https://openaccess.thecvf.com/content/WACV2025/papers/Karri_Uncertainty-Guided_Cross_Attention_Ensemble_Mean_Teacher_for_Semi-Supervised_Medical_Image_WACV_2025_paper.pdf}
}

@inproceedings{tan2025dpa,
  title     = {Decoupled Proxy Alignment: Mitigating Language Prior Conflict for Multimodal Alignment in MLLM},
  author    = {Chenkun Tan and Pengyu Wang and Shaojun Zhou and Botian Jiang and Zhaowei Li and Dong Zhang and Xinghao Wang and Yaqian Zhou and Xipeng Qiu},
  booktitle = {Findings of the Association for Computational Linguistics: EMNLP 2025},
  year      = {2025},
  note      = {arXiv:2509.14735},
  url       = {https://arxiv.org/abs/2509.14735}
}

@article{cohn2024mmla_review,
  title   = {Multimodal Methods for Analyzing Learning and Training Environments: A Systematic Literature Review},
  author  = {Clayton Cohn and Eduardo Davalos and Caleb Vatral and Joyce Horn Fonteles and Hanchen David Wang and Meiyi Ma and Gautam Biswas},
  journal = {arXiv preprint arXiv:2408.14491},
  year    = {2024},
  url     = {https://arxiv.org/pdf/2408.14491}
}

@inproceedings{sun2024redcore,
  title     = {RedCore: Relative Advantage Aware Cross-modal Representation Learning for Missing Modalities with Imbalanced Missing Rates},
  author    = {Jun Sun and Xinxin Zhang and Shoukang Han and Yu-Ping Ruan and Taihao Li},
  booktitle = {Proceedings of the AAAI Conference on Artificial Intelligence (AAAI)},
  year      = {2024},
  pages     = {to appear},
  url       = {https://arxiv.org/abs/2312.10386}
}

@article{cai2019mmsd,
  title   = {Multi-Modal Sarcasm Detection in Twitter with Hierarchical Fusion Model},
  author  = {Cai, Yitao and Cai, Huiyu and Wan, Xiaojun},
  journal = {Proceedings of the 57th Annual Meeting of the Association for Computational Linguistics},
  year    = {2019}
}

@inproceedings{Sun2024CAGC,  
  author    = {Kaili Sun and Zhiwen Xie and Mang Ye and Huyin Zhang},  
  title     = {Contextual Augmented Global Contrast for Multimodal Intent Recognition},  
  booktitle = CVPR,  
  pages     = {26963--26973},  
  year      = {2024},  
  doi       = {10.1109/CVPR52733.2024.02546}  
}

@inproceedings{Wei2024G2SAM,
  author    = {Yiwei Wei and Shaozu Yuan and Hengyang Zhou and Longbiao Wang and Zhiling Yan and Ruosong Yang and Meng Chen},
  title     = {G²SAM: Graph-Based Global Semantic Awareness Method for Multimodal Sarcasm Detection},
  booktitle = {Proc.\ of the 38th Annual AAAI Conference on Artificial Intelligence (AAAI 2024) — Technical Tracks},
  pages     = {9151--9159},
  year      = {2024},
  doi       = {10.1609/AAAI.v38i8.28766}
}

@inproceedings{Devlin2019BERT,
  author    = {Jacob Devlin and Ming-Wei Chang and Kenton Lee and Kristina Toutanova},
  title     = {BERT: Pre-training of Deep Bidirectional Transformers for Language Understanding},
  booktitle = {Proceedings of the 2019 Conference of the North American Chapter of the Association for Computational Linguistics: Human Language Technologies (NAACL-HLT)},
  volume    = {1 (Long and Short Papers)},
  pages     = {4171--4186},
  year      = {2019},
  doi       = {10.18653/v1/N19-1423}
}

@inproceedings{Baltrusaitis2016OpenFace,
  author    = {Tadas Baltrušaitis and Peter Robinson and Louis-Philippe Morency},
  title     = {OpenFace: an open source facial behavior analysis toolkit},
  booktitle = {Proceedings of the 2016 IEEE Winter Conference on Applications of Computer Vision (WACV)},
  pages     = {1—10},
  year      = {2016}
}

@inproceedings{Degottex2014COVAREP,
  author    = {Gilles Degottex and John Kane and Thomas Drugman and Tuomo Raitio and Stefan Scherer},
  title     = {COVAREP – A collaborative voice analysis repository for speech technologies},
  booktitle = {Proceedings of the IEEE International Conference on Acoustics, Speech and Signal Processing (ICASSP)},
  address   = {Florence, Italy},
  year      = {2014}
}

@inproceedings{Williams2018RecEmoVideo,
  author    = {Jennifer Williams and Steven Kleinegesse and Ramona Comanescu and Oana Radu},
  title     = {Recognizing Emotions in Video Using Multimodal DNN Feature Fusion},
  booktitle = {Proceedings of the First Grand Challenge and Workshop on Human Multimodal Language (Challenge-HML)},
  pages     = {11--19},
  year      = {2018},
  doi       = {10.18653/v1/W18-3302}
}

@inproceedings{Pham2019FoundInTranslation,
  author    = {Hai Pham and Truyen Tran and Svetha Venkatesh},
  title     = {Found in Translation: Learning Robust Joint Representations by Cyclic Translations Between Modalities},
  booktitle = {Proceedings of the AAAI Conference on Artificial Intelligence (AAAI)},
  volume    = {33},
  number    = {01},
  pages     = {6892--6899},
  year      = {2019},
  doi       = {10.1609/aaai.v33i01.33016892}
}

@inproceedings{Yu2021SelfMM,
  author    = {Wenmeng Yu and Hua Xu and Ziqi Yuan and Jiele Wu},
  title     = {Learning Modality-Specific Representations with Self-Supervised Multi-Task Learning for Multimodal Sentiment Analysis},
  booktitle = {Proceedings of the AAAI Conference on Artificial Intelligence},
  volume    = {35},
  number    = {12},
  pages     = {10790--10797},
  year      = {2021},
  doi       = {10.1609/aaai.v35i12.17289}
}

@inproceedings{Han2021MMIM,
  author    = {Wei Han and Hui Chen and Soujanya Poria},
  title     = {Improving Multimodal Fusion with Hierarchical Mutual Information Maximization for Multimodal Sentiment Analysis},
  booktitle = {Proceedings of the 2021 Conference on Empirical Methods in Natural Language Processing (EMNLP)},
  pages     = {9180--9192},
  year      = {2021},
  doi       = {10.18653/v1/2021.emnlp-main.723}
}

@inproceedings{Rahman2020MAG,
  author    = {Wasifur Rahman and Md Kamrul Hasan and Sangwu Lee and AmirAli Bagher Zadeh and Chengfeng Mao and Louis{-}Philippe Morency and Ehsan Hoque},
  title     = {Integrating Multimodal Information in Large Pretrained Transformers},
  booktitle = {Proceedings of the 58th Annual Meeting of the Association for Computational Linguistics},
  pages     = {2359--2369},
  year      = {2020},
  doi       = {10.18653/v1/2020.acl-main.214}
}

@inproceedings{zhu2025emosym,
  title={Emosym: A symbiotic framework for unified emotional understanding and generation via latent reasoning},
  author={Zhu, Yijie and Lyu, Yibo and Yu, Zitong and Shao, Rui and Zhou, Kaiyang and Nie, Liqiang},
  booktitle={Proceedings of the 33rd ACM International Conference on Multimedia},
  pages={5451--5460},
  year={2025}
}

@inproceedings{zhu2026h,
  title={H-gar: A hierarchical interaction framework via goal-driven observation-action refinement for robotic manipulation},
  author={Zhu, Yijie and Shao, Rui and Liu, Ziyang and He, Jie and Liu, Jizhihui and Wang, Jiuru and Yu, Zitong},
  booktitle={Proceedings of the AAAI Conference on Artificial Intelligence},
  volume={40},
  number={22},
  pages={18882--18890},
  year={2026}
}

@article{zhu2026uniemo,
  title={Uniemo: Unifying emotional understanding and generation with learnable expert queries},
  author={Zhu, Yijie and Zhang, Lingsen and Yu, Zitong and Shao, Rui and Tan, Tao and Nie, Liqiang},
  journal={IEEE Transactions on Image Processing},
  year={2026},
  publisher={IEEE}
}

@article{zhu2026delta,
  title={{$\Delta$-VLA}: Prior-Guided Vision-Language-Action Models via World Knowledge Variation},
  author={Zhu, Yijie and He, Jie and Shao, Rui and Yuan, Kaishen and Tan, Tao and Yuan, Xiaochen and Yu, Zitong},
  journal={arXiv preprint arXiv:2603.08361},
  year={2026}
}

@article{zhu2026ati,
  title={ATI-VLA: Action-Centric Predictive Vision-Language-Action Models via Actionable Alignment Then Adaptive Injection},
  author={Zhu, Yijie and Shao, Rui and He, Jie and Li, Wei and Zhao, Bo and Wang, Yelin and Yuan, Xiaochen and Tan, Tao and Zhang, Miao and Peng, Xiaojiang and others},
  journal={arXiv preprint arXiv:2610.01741},
  year={2026}
}

@article{zhu2026learning,
  title={Learning to Use Imagination: Progress-Conditioned Future Utilization for World Action Models},
  author={Zhu, Yijie and Yu, Zitong and Li, Wei and Ma, Hui and Li, Wen and Shao, Rui and Nie, Liqiang},
  journal={arXiv preprint arXiv:2609.06578},
  year={2026}
}

@article{wang2026affectagent,
  title={AffectAgent: Collaborative Multi-Agent Reasoning for Retrieval-Augmented Multimodal Emotion Recognition},
  author={Wang, Zeheng and Yu, Zitong and Zhu, Yijie and Zhao, Bo and Liang, Haochen and Wang, Taorui and Xia, Wei and Zhang, Jiayu and Liu, Zhishu and Ma, Hui and others},
  journal={arXiv preprint arXiv:2604.12735},
  year={2026}
}

@article{wang2026navigating,
  title={Navigating the Emotion Tree: Hierarchical Hyperbolic RAG for Multimodal Emotion Recognition},
  author={Wang, Zeheng and Zhao, Bo and Zhu, Yijie and Liu, Zhishu and Ma, Hui and Zhang, Ruixin and Ding, Shouhong and Xie, Qianyu and Yu, Zitong},
  journal={arXiv preprint arXiv:2605.18884},
  year={2026}
}

@misc{gao2026polarityasymmetricstructuralcalibrationlink,
  title={Polarity-Asymmetric Structural Calibration for Link Sign Prediction},
  author={Qiqi Gao and Wenzhuo Song and Xueyan Liu},
  year={2026},
  eprint={2609.05896},
  archivePrefix={arXiv},
  primaryClass={cs.SI},
  url={https://arxiv.org/abs/2609.05896}
}

@misc{wang2026signedgraph,
  author        = {Wang, Zifan and Chen, Siyu and Song, Wenzhuo},
  title         = {Signed-Graph Recommendation as Structural Consistency Maximization},
  year          = {2026},
  eprint        = {2607.05952},
  archivePrefix = {arXiv},
  primaryClass  = {cs.SI},
  doi           = {10.48550/arXiv.2607.05952}
}

@inproceedings{wang2027manchutts,
  author    = {Wang, Suhua and Wang, Zifan and Sun, Xiaoxin and Wang, Meiqi and Zhang, Yun and Li, Xin},
  title     = {{ManchuTTS}: Towards High-Quality Manchu Speech Synthesis via Flow Matching and Hierarchical Text Representation},
  booktitle = {Advanced Intelligent Computing Technology and Applications},
  editor    = {Huang, De-Shuang and Li, Bo and Zhang, Qinhu and Bao, Wenzheng},
  series    = {Lecture Notes in Computer Science},
  volume    = {16669},
  pages     = {51--61},
  publisher = {Springer Nature Singapore},
  address   = {Singapore},
  year      = {2027},
  doi       = {10.1007/978-981-92-3492-9_5}
}

@article{ma2026cm,
  title={CM-CLIP: few-shot cross-modal generalization for face anti-spoofing via prompt learning},
  author={Ma, Hui and Liu, Ajian and Lin, Xun and Escalante, Hugo Jair and Guyon, Isabelle and Wan, Jun and Lei, Zhen and Liang, Yanyan},
  journal={Machine Intelligence Research},
  volume={23},
  number={2},
  pages={366--382},
  year={2026},
  publisher={Springer}
}

@article{ma2026clip,
  title={CLIP-SA: CLIP-Guided Semantic Alignment for Generalizable Face Anti-spoofing},
  author={Ma, Hui and Liu, Ajian and Xue, Shuai and Diao, Yunfeng and Liang, Dashuang and Liang, Hailun and Liang, Yanyan and Wan, Jun and Yang, Xun and Hu, ZhenZhen and others},
  journal={IEEE Transactions on Multimedia},
  year={2026},
  publisher={IEEE}
}

@article{ma2026flexible,
  title={Flexible Modal Mixture-of-Experts with Inter-Modal Knowledge Distillation for Face Anti-Spoofing},
  author={Ma, Hui and Liu, Ajian and Li, Ning and Yang, Liying and Wang, Boyun and Zou, Hang and Zhang, Yuan and Huang, Jing and Pu, Zhiqiang and Wan, Jun and others},
  journal={IEEE Transactions on Information Forensics and Security},
  year={2026},
  publisher={IEEE}
}

@article{liu2026aullm++,
  title={AULLM++: Structural Reasoning with Large Language Models for Micro-Expression Recognition},
  author={Liu, Zhishu and Yuan, Kaishen and Zhao, Bo and Ma, Hui and Yu, Zitong},
  journal={arXiv preprint arXiv:2603.08387},
  year={2026}
}

@article{wang2026micro,
  title={Micro-gesture recognition: A comprehensive survey of datasets, methods, and challenges},
  author={Wang, Taorui and Lin, Xun and Xu, Yong and Ye, Qilang and Guo, Dan and Escalera, Sergio and Khoriba, Ghada and Yu, Zitong},
  journal={Machine Intelligence Research},
  volume={23},
  number={2},
  pages={308--330},
  year={2026},
  publisher={Springer}
}

@inproceedings{10.1145/3767308.3835254,
  author = {Wang, Taorui and Xia, Wei and Ma, Hui and Song, Zijia and Zhang, Jiayu and Wang, Zeheng and Xu, Yong and Yu, Zitong},
  title = {GMoT: Gated Motion-Aware Tokenization for Fine-Grained Micro-Gesture Video Reasoning with Multimodal LLMs},
  year = {2026},
  isbn = {9798400722134},
  publisher = {Association for Computing Machinery},
  address = {New York, NY, USA},
  url = {https://doi.org/10.1145/3767308.3835254},
  doi = {10.1145/3767308.3835254},
  booktitle = {Proceedings of the 34th ACM International Conference on Multimedia},
  pages = {10596--10605},
  numpages = {10},
  location = {Rio de Janeiro, Brazil},
  series = {MM '26}
}

@inproceedings{10.1007/978-981-92-4529-1_3,
  author = {Wang, Taorui and Ma, Hui and Xia, Wei and Zhang, Jiayu and Huang, Jiajian and Xu, Yong},
  editor = {Chen, Haoyu and Ren, Zhao and Wang, Jiadong and Provost, Emily and Adeli, Ehsan and Schuller, Bj{\"o}rn W. and Schultz, Tanja and Zhao, Guoying},
  title = {SwinT-Enhanced Class-Wise Fusion for Micro-gesture Classification},
  booktitle = {Human Behavior Analysis and Interaction for Emotional Intelligence},
  year = {2027},
  publisher = {Springer Nature Singapore},
  address = {Singapore},
  pages = {28--34},
  isbn = {978-981-92-4529-1}
}

@article{11658652,
  author={Wang, Yelin and Zhu, Yijie and Song, Zijia and Ma, Hui and Wang, Zeheng and Yu, Zitong and Ma, Fei and Tian, Qi},
  journal={IEEE Transactions on Affective Computing},
  title={Uni-EmoAgent: A Unified Multi-Agent Framework for Affective Understanding and Art-Oriented Emotional Image Generation},
  year={2026},
  pages={1--13},
  doi={10.1109/TAFFC.2026.3725184}
}

@inproceedings{wang2026rsiccllm,
  title     = {{RSICCLLM}: A Multimodal Large Language Model for Remote Sensing Image Change Captioning},
  author    = {Wang, Yelin and Song, Zijia and Ye, Shuo and Yang, Chuanguang and Wang, Miaoyu and Xu, Yong and An, Zhulin and Xu, Yongjun and Yu, Zitong},
  booktitle = {Computer Vision -- ECCV 2026},
  series    = {Lecture Notes in Computer Science},
  volume    = {17056},
  pages     = {253--272},
  publisher = {Springer},
  year      = {2026},
  doi       = {10.1007/978-3-032-37324-3_14}
}

@inproceedings{wang2026dfm,
  title     = {{DFM}: Difference Feature Modeling with Text-Guided Gated Contrastive Loss for Remote Sensing Image Change Captioning},
  author    = {Wang, Yelin and Song, Zijia and Yang, Chuanguang and Wang, Miaoyu and An, Zhulin and Huang, Libo and Xu, Yongjun},
  booktitle = {2026 IEEE International Conference on Multimedia and Expo (ICME)},
  year      = {2026},
  url       = {https://arxiv.org/abs/2606.27410}
}

@inproceedings{tan2026veritas,
  title={Veritas: Generalizable deepfake detection via pattern-aware reasoning},
  author={Tan, Hao and Tan, Zichang and Shi, Senyuan and Liu, Ajian and Song, Chuanbiao and Zhu, Huijia and Wang, Weiqiang and Wan, Jun and Lei, Zhen and others},
  booktitle={International Conference on Learning Representations},
  volume={2026},
  pages={66420--66477},
  year={2026}
}

@article{tan2026videoveritas,
  title={Videoveritas: Ai-generated video detection via perception pretext reinforcement learning},
  author={Tan, Hao and Lan, Jun and Shi, Senyuan and Tan, Zichang and Yu, Zijian and Zhu, Huijia and Wang, Weiqiang and Wan, Jun and Lei, Zhen},
  journal={arXiv preprint arXiv:2602.08828},
  year={2026}
}

@article{shi2026hydraprompt,
  title={HydraPrompt: An Adaptive and Asymmetric Framework of Vision-Language Models for Synthetic Image Detection},
  author={Shi, Senyuan and Tan, Hao and Tan, Zichang and Feng, Shuhan and Liu, Ajian and Escalera, Sergio and Wan, Jun},
  journal={arXiv preprint arXiv:2605.26421},
  year={2026}
}

@inproceedings{DBLP:conf/iclr/PanCWDW26,
  author       = {Yu Pan and Jiahao Chen and Lin Wang and Bingrong Dai and Wenjie Wang},
  editor       = {Carl Vondrick and Bharath Hariharan and Colin Raffel and Lerrel Pinto and Diyi Yang and Aleksandra Faust},
  title        = {STEDiff: Revealing the Spatial and Temporal Redundancy of Backdoor Attacks in Text-to-Image Diffusion Models},
  booktitle    = {The Fourteenth International Conference on Learning Representations, {ICLR} 2026, Rio de Janeiro, Brazil, April 23-27, 2026},
  publisher    = {proceedings.iclr.cc},
  year         = {2026},
  url          = {https://proceedings.iclr.cc/paper\_files/paper/2026/hash/559726fdfb19005e368be4ce3d40e3e5-Abstract-Conference.html}
}

@article{DBLP:journals/corr/abs-2605-07399,
  author       = {Yu Pan and Andi Zhang and Yi Wang and Sibei Yang and Wenjie Wang},
  title        = {{GPO-V:} Jailbreak Diffusion Vision Language Model by Global Probability Optimization},
  journal      = {CoRR},
  volume       = {abs/2605.07399},
  year         = {2026},
  url          = {https://doi.org/10.48550/arXiv.2605.07399},
  doi          = {10.48550/ARXIV.2605.07399},
  eprinttype   = {arXiv},
  eprint       = {2605.07399}
}

\end{document}